\documentclass[journal]{IEEEtran}
\usepackage{amsmath,amsfonts}
\usepackage{algpseudocode}
\usepackage[linesnumbered,ruled,vlined]{algorithm2e}
\usepackage{array}
\usepackage[caption=false,font=normalsize,labelfont=sf,textfont=sf]{subfig}
\usepackage{textcomp}
\usepackage{stfloats}
\usepackage{url}
\usepackage{verbatim}
\usepackage{graphicx}
\usepackage{cite}
\usepackage{amssymb}
\usepackage[mathscr]{eucal} 
\usepackage{diagbox}
\usepackage{makecell}
\usepackage{xcolor}
\usepackage{units}
\usepackage{multirow}

\newcommand{\ie}{\textit{i.e.,}\ }
\newcommand{\eg}{\textit{e.g.,}\ }

\DeclareMathOperator*{\argmin}{arg\,min}

\newcommand{\bigO}{\mathcal{O}}

\newtheorem{definition}{Definition}

\newtheorem{remark}{Remark}

\newtheorem{problem}{Problem}

\newcommand{\edit}[1]{\textcolor{black}{#1}}

\begin{document}

\title{Distributed Multi-Robot Multi-Target Tracking Using Heterogeneous Limited-Range Sensors}


\author{Jun Chen,~\IEEEmembership{Member,~IEEE}, Mohammed Abugurain,~\IEEEmembership{Student Member,~IEEE}, Philip Dames,~\IEEEmembership{Member,~IEEE}, and Shinkyu Park,~\IEEEmembership{Member,~IEEE}
\thanks{*This work was supported by startup funding at Nanjing Normal University, funding from King Abdullah University of Science and Technology, and NSF grant CNS-2143312.}
\thanks{J.~Chen is with the School of Electrical and Automation Engineering, Nanjing Normal University, Nanjing, Jiangsu 210023, China
        {\tt\small jun.chen@nnu.edu.cn}}
\thanks{\edit{J.~Chen is also with Jiangsu Key Laboratory of 3D Printing Equipment and Manufacturing, Nanjing, Jiangsu 210023, China}}
\thanks{M.~Abugurain and S.~Park are with the Computer, Electrical, and Mathematical Science and Engineering Division, King Abdullah University of Science and Technology, Thuwal 23955, Saudi Arabia
        {\tt\small \{mohammed.abugurain,shinkyu.park\}
        @kaust.edu.sa}}
\thanks{P.~Dames is with the Department of Mechanical Engineering, Temple University, Philadelphia, PA 19122, USA
        {\tt\small pdames@temple.edu}}%
}



\maketitle

\begin{abstract}
This paper presents a cooperative multi-robot multi-target tracking framework aimed at enhancing the efficiency of the heterogeneous sensor network and, consequently, improving overall target tracking accuracy. The concept of \emph{normalized unused sensing capacity} is introduced to quantify the information a sensor is currently gathering relative to its theoretical maximum. This measurement can be computed using entirely local information and is applicable to various sensor models, distinguishing it from previous literature on the subject. It is then utilized to develop a distributed coverage control strategy for a heterogeneous sensor network, adaptively balancing the workload based on each sensor's current unused capacity. The algorithm is validated through a series of ROS and \textsc{MATLAB} simulations, demonstrating superior results compared to standard approaches that do not account for heterogeneity or current usage rates.
\end{abstract}


\section{Introduction}
\label{sec:intro}
\IEEEPARstart{M}{ultiple} target tracking (MTT) is a fundamental research problem where one needs to continuously estimate the states of multiple moving targets of interest within an assigned space. 
Due to its importance in application areas, ranging from environmental monitoring, e.g., comprehending collective behaviours of social animals \cite{macgregor2020information, delellis2014collective, gomez2022intermittent,chuang2016underwater} or pedestrians \cite{scovanner2009learning}, to intelligent autonomous systems, e.g., autonomous driving \cite{adnan2023comprehensive}, it has drawn increasing attention in the signal processing, computer vision, and robotics communities.

Standard MTT algorithms, such as global nearest neighbor (GNN) \cite{konstantinova2003study}, joint probabilistic data association (JPDA) \cite{hamid2015joint}, multiple hypothesis tracking (MHT) \cite{blackman2004multiple}, and particle filters \cite{doucet2002particle}, solve the problem by repeatedly identifying targets and estimating their states using sensor data and Bayesian inferences, and matching the estimates to existing target tracking information, the process widely known as data association.
On the other hand, solutions that are based on random finite set (RFS) theory, such as the family of probability hypothesis density (PHD) filters \cite{mahler2003multitarget,mahler2007phd} and multi-target multi-Bernoulli (MeMBer) filters \cite{mahler2007statistical,vo2009cardinality}, do not require an explicit data association process. 
Hence, those approaches are best suited for the applications where the identities of targets are not important.

\edit{\subsection{Distributed MR-MTT}}
In recent years, exploring robotic networks' ability to adaptively adjust their sensor coverage, multi-robot multi-target tracking (MR-MTT) problems have been widely investigated.
In this paper, we discuss distributed algorithms to find scalable and fault-tolerant approaches to the problems.
Distributed MR-MTT mainly consists of cooperative multi-target state estimation and multi-robot active motion planning.
The former can be implemented using either Bayesian methods \cite{olfati2007distributed, campbell2016distributed, hollinger2014distributed, zhu2020fully, khodayi2019distributed}, \edit{which require each robot to recursively estimate the statistical distribution about target states over the entire task space}, or consensus algorithms \cite{ge2019distributed, olfati2005consensus, shirsat2022probabilistic}, which enable robots to continuously exchange information until they reach an agreement. 
The latter is frequently formulated as an optimal control problem aiming to \edit{maximize the tracking performance, which is defined as} the degree of information acquisition \cite{corah2021scalable, schlotfeldt2021resilient, julian2012distributed}, the distance between a robot and a target \cite{sung2020distributed}, \edit{or} some general measure of the target observability \cite{zhou2019sensor, khodayi2019distributed}.
However, the above-mentioned schemes could be susceptible to local \edit{extrema}, where multiple robots track the same target while others are left unobserved, and provide no guarantee to cover the entire task space with the robots' limited-range sensors.
\edit{As a result, none of them is able to jointly detect, localize, and track an unknown and time varying number of multiple targets.}

\edit{\subsection{Coverage Control}}
\edit{Perhaps the most suitable MR-MTT solution to pair space exploration with active target state estimation is through coverage control framework.}
Coverage control is a popular way to optimally gather target information while guaranteeing complete space coverage.
It aims to maximize a performance measure such as the robots' distances to targets.
The most widely used distributed coverage control strategies are Voronoi-based methods \cite{cortes2004coverage, schwager2009decentralized, kantaros2016distributed, pierson2017distributed} that leverage Lloyd's algorithm \cite{lloyd1982least}. 
The basic idea is to partition an assigned task space using the Voronoi tessellation and then drive each robot towards the weighted centroid of its assigned Voronoi partition.
This process will enable the network to reach a local optimum in its coverage performance.

In coverage control, the mission space is typically convex, and \edit{robots} are only required to communicate with their neighbors and are assumed to have perfect knowledge of their locations.
A number of recent works have provided various solutions to accommodate uncertainty in sensor localization \cite{papatheodorou2018distributed, zhu2019b, wang2019distributed, chen2020distributed} and to allow for non-convex spaces \cite{palacios2017optimal, bhattacharya2014multi, breitenmoser2010voronoi, pimenta2008sensing}. 
\emph{In particular, leveraging the coverage control framework, the work of Dames \cite{dames2020distributed} proposes to couple a distributed multi-target state estimation scheme and a distributed planning algorithm in order for each robot to estimate the density of targets and move to the centroid of the estimated target density.}
Consequently, networked robots with a limited-range isotropic sensor are able to actively move to maximize the total detection probability of all targets despite measurement errors and false alarms, while maintaining coverage of the entire task space.

\edit{\subsection{Sensor Heterogeneity}}
\edit{Nevertheless, the above-mentioned work exploits MR-MTT with coverage control framework using fixed types of sensor.
Yet, }cooperative heterogeneous sensor networks \edit{composed of arbitrary types of sensors} expand the use of robotic networks to complex scenarios where a variety of different types of robots and sensors are needed to complete the task.
A variety of approaches have been proposed to incorporate measurements from multiple heterogeneous resources to improve the target tracking quality \cite{kushwaha2008multi, helgesen2022heterogeneous, zhou2008target} or to enhance the robustness and resilience of a robotic network \cite{mayya2022adaptive, ramachandran2019resilience}. 
However, none of these methods aims to attain the globally optimal tracking performance by taking advantage of heterogeneous sensing capability while maintaining coverage of the task space.

Weighted Voronoi diagrams, which include power diagrams, have recently been used to account for heterogeneity in the radii of circular sensor footprints \cite{pimenta2008sensing, kantaros2015distributed, bartolini2010autonomous, arslan2016voronoi} and energy levels \cite{kwok2010deployment}.
Voronoi-based methods have also been used for wedge-shaped sensor field of views (FoV) \cite{laventall2008coverage}. 
Stergiopoulos and Tzes \cite{stergiopoulos2012autonomous, stergiopoulos2013spatially, stergiopoulos2014cooperative} used novel partitioning and distribution algorithms for sensing networks where all sensor FoVs share the same arbitrary shape but may be of a different size.
Hexsel \textit{et al.}~\cite{hexsel2011coverage} introduced a gradient ascent-based coverage control algorithm to maximize the joint probability of detection in anisotropic sensing networks.
Nevertheless, as these approaches use gradient ascent, robots can prematurely converge to local extrema of an objective function they aim to maximize.

\edit{\subsection{Contribution}}
Building on our previous conference version \cite{chen2021distributed}, this paper proposes a novel multi-robot multi-target joint state estimation and planning approach for heterogeneous limited-FoV robotic networks. 
\edit{The proposed method allows a team of robots to search for targets over a task space and actively maintain coverage of a majority of detected targets in a distributed manner.}
We summarize our main contributions as follows.
\begin{enumerate}
  \item We propose a new measure called the \emph{normalized unused sensing capacity} that quantifies the difference between the current information that a sensor gathers and the theoretical maximum.
  This can be computed using entirely local information and does not require any assumptions about the type of sensor or the shape of the sensor FoV.
  \item \edit{Leveraging this measure, we first replace the standard Voronoi diagram in Lloyd's algorithm with the power diagram, with the goal of balancing the task load across the team.}
  In particular, we assign robots that have more accurate sensors, larger sensor FoVs, and/or are not currently tracking any objects to cover larger areas.
  \edit{\item While power diagram implementation provides fast and near optimal space allocation for heterogeneous robotic teams, it does not yield the best tracking accuracy. Therefore, we propose CCVD, a closed-form optimal space partitioning algorithm to further improve the tracking performance and to evaluate the power diagram method by comparing it with theoretical optimum.}  
  \item We demonstrate the efficacy of our approach through a series of experiments in simulated environments to show that the approach yields higher quality and more reliable tracking \edit{than the standard Voronoi diagram-based approach and the zigzag coverage path}, especially when the robots are highly heterogeneous. \edit{To the best of our knowledge, our work presents the first distributed algorithm that allows heterogeneous sensors to jointly detect, localize and track unknown and time-varying number of targets. Therefore, our results are only comparable to baseline algorithms \cite{dames2020distributed}\cite{torres2016coverage}.}
\end{enumerate}
Distinct from our conference paper \cite{chen2021distributed}, this paper includes the following novel contributions.
\begin{enumerate}
  \item Whereas only the power diagram was used in \cite{chen2021distributed}, in this paper, we introduce a new space partitioning algorithm based on CCVD. 
  \edit{The area of the task space allocated to each robot by CCVD is strictly proportional to the sensing capacity of that robot, which will further optimize the space partitioning by leveraging the heterogeneity of the robots, but at the cost of greater computing resource requirements.}
  \item We analyze the computational complexity of the proposed algorithms, and discuss the communication cost incurred by the algorithms.
  \item We present ROS simulations with visualization in Gazebo and RViz. We conduct additional \textsc{Matlab} simulations for quantitative analysis of the new methods.
  \item \edit{We compare our methods with a coverage path planning approach quantitatively. Meanwhile, we conduct an ablation study to demonstrate the usage of both the centroid of detection and the partitioning algorithms.}
\end{enumerate}

The rest of this paper is organized as follows. 
In Section~\ref{sec:problem}, we formulate the MR-MTT problem and introduce the prerequisites of our proposed methods.
Section~\ref{sec:partition} presents our proposed space partitioning algorithms using both the power diagram and CCVD along with a distributed control algorithm based on these partitions.
In Section~\ref{sec:performance} we analyze the performance of our algorithms in terms of computational complexity and communication \edit{loac} and introduce the quantitative metrics used to validate our algorithms.
Finally, we show simulation results in Section~\ref{sec:results} and draw conclusions in Section~\ref{sec:conclusions}.

\section{Problem Formulation}
\label{sec:problem}

\edit{
This paper considers the multi-robot, multi-target tracking (MR-MTT) problem, defined below.
\begin{problem}[MR-MTT]
Consider a network of $n$ mobile sensors (\ie robots) denoted by $S = \{ s_1, \ldots, s_n \}$ with two-dimensional positions $Q = \{q_1, \ldots, q_n \}$ and orientations $\Theta = \{\theta_1, \ldots, \theta_n \}$.
Each robot has the following dynamical model:
\begin{equation}
\begin{aligned}
\dot{q_i} &= u_i\\
\dot{\theta_i} &= \omega_i,
\label{eq:dynamic}
\end{aligned}    
\end{equation}
where $u_i$ and $\omega_i$ are the two-dimensional control inputs for position and orientation of the $i$th robot, respectively.
The robots move in a convex environment $E \subset R^2$ with known bounds.
There is a set of $m$ targets in the environments, $X = \{x_1, \ldots, x_m\}$, where the state of each target is its two-dimensional position $x_i \in E$.
\edit{Targets may be moving in an unknown arbitrary pattern or be stationary, and the robot's onboard sensors have a certain probability of detecting them.}
The robots are tasked to search for and track the targets, with the goal to identify the number of targets and the state of each target.
\end{problem}
}

\edit{
Robots communicate with each other bidirectionally. 
A neighbor set $\mathcal{N}_i$ of a robot $s_i$ is defined as all robots that are within the communication range of robot $s_i$ excluding $s_i$ itself.
For the purpose of analysis, we assume that the communication range of each robot is large enough such that $\mathcal N_i$ is non-empty at all time, which can be ensured by using, 
for instance, connectivity control algorithms \cite{luo2019voronoi} to sustain the connectivity.\footnote{Existing methods such as \cite{palacios2017optimal, bhattacharya2014multi, breitenmoser2010voronoi, pimenta2008sensing} can be used to extend our proposed algorithms to non-convex, cluttered environments.}
}


\subsection{Lloyd's Algorithm}
Given a density function $\phi(x)$ defined over the 2-dimensional Euclidean space, the objective of Lloyd's algorithm is to compute $Q = {q_1, \cdots, q_n}$ and $\mathcal W = (\mathcal W_1, \cdots, \mathcal W_n)$ minimizing the following functional:
\begin{equation}
\mathcal{H}(Q, \mathcal{W}) = \sum_{i=1}^n \int_{\mathcal{W}_i} f\big(\|x - q_i \|\big) \phi(x) dx,
\label{eq:optimization}
\end{equation}
where $\mathcal{W}_i$ is the dominance region of robot $s_i$ (\textit{i.e.}, the region that is assigned to robot $s_i$ for coverage), $\| \cdot \|$ is the Euclidean distance, $x \in E$, and $f$ is a monotonically increasing function. 
The function $f$ defines how the cost of robots' sensing depends on its distance to each sensing location $x$.
The dominance regions $\mathcal{W}$ partition $E$, meaning the regions have disjoint interiors (\ie $\mathrm{int}(\mathcal{W}_i) \cap \mathrm{int}(\mathcal{W}_j) = \emptyset \; \forall i \neq j$) and the union of all regions is $E$ (\ie $\cup_{i=1}^n \mathcal{W}_i = E$) \cite{cortes2004coverage}.

When $f(x) = x^2$, each $\mathcal W_i$ of $\mathcal{W}$ is given by $\mathcal W_i = \{x \mid \|x - q_i\| = \min_{k = 1, \ldots, n} \|x - q_k\| \}$ \cite{cortes2004coverage}.
In other words, $\mathcal W_i$ is the collection of all points that are the nearest neighbors of $s_i$.
This is widely known as the Voronoi tessellation, which is illustrated in Figure \ref{fig:diagrams}.
We refer to $\mathcal W_i$ as a Voronoi partition, which is convex by construction, and $q_i$ as the generation point of $\mathcal W_i$.
When each $q_i$ is the weighted centroid of the $i$-th Voronoi partition given by
\begin{equation}
q_i^{\ast} = \frac{\int_{\mathcal W_i} x \phi(x) \, dx}{\int_{\mathcal W_i} \phi(x) \, dx},
\label{eq:lloyd's algorithm}
\end{equation}
the robot positions $Q$ minimize $\mathcal{H}$ \cite{cortes2004coverage}.

By applying Lloyd's algorithm, coverage control sets the control input for robot $s_i$ to
\begin{equation}
u_i = -k_{\rm prop}(q_i - q_i^{\ast}),
\label{eq:control law}
\end{equation}
where $k_{\rm prop} > 0$ is a positive gain.
In this paper, we use the control law in a discrete time manner, following this direction at the maximum speed of the robot.
The angular velocity uses a bang-bang strategy, maximizing the angular velocity until the robot is facing directly towards the goal~$q_i^\ast$.
By following this control law, the robots asymptotically converge to the weighted centroids of their associated Voronoi partitions.
This still holds even when $\phi$ varies with time.

\subsection{PHD Filter for MTT}
\label{subsec:phd}
In an MTT setting, a natural choice for $\phi (x)$ is to capture the target density at each location $x$.
This time-varying density function can be estimated using any standard MTT algorithms.
In this paper, we use the PHD filter, as it does not require any explicit data association.\footnote{We note that the overall framework presented in this paper can be easily adapted to any other choice of Bayesian MTT tracker.}

The \emph{Probability Hypothesis Density} (PHD), denoted by $v(x)$, is the first order moment of a random finite set (RFS), \ie a set with a random number of random elements such as a time-varying set of moving targets, and takes the form of a density function over the state space.
The integral of the PHD over a region is equal to the expected number of targets in the region at a given time. 
By assuming that the RFSs are Poisson, \ie the number of targets follows a Poisson distribution and the spatial distribution of targets is i.i.d., the PHD filter recursively propagates the PHD in order to track the distribution over target sets \cite{mahler2003multitarget}.
Similar to any other Bayesian state estimators, the PHD filter recursively predicts the target state using state transition probability (\ie target motion), and updates the prediction using the probabilistic models of the sensor measurements.

The PHD filter uses three models to describe the motion of targets:
1) The motion model, $f(x \mid \xi)$, describes the likelihood of an individual target transitioning from an initial state $\xi$ to a new state $x$.
2) The survival probability model, $p_s(x)$, describes the likelihood that a target with state $x$ will continue to exist from one time step to the next.
3) The birth PHD, $b(x)$, encodes both the number and locations of the new targets that may appear in the environment.

The PHD filter uses another three models to describe the ability of robots to detect targets:
1) The detection model, $p_d(x \mid q)$, gives the probability of a robot with state $q$ successfully detecting a target with state $x$.
For instance, we can set the detection probability to zero for all $x$ outside the sensor FoV.
2) The measurement model, $g(z \mid x, \, q)$, gives the likelihood of a robot with state $q$ receiving a measurement $z$ from a target with state $x$.
3) The false positive (or clutter) PHD, $c(z \mid q)$, describes both the number and locations of the clutter measurements.

Using these target and sensor models, the PHD filter prediction \edit{\eqref{eq:phd_prediction}} and update \edit{\eqref{eq:phd_update}} equations are as follows:
\begin{subequations}
\begin{align}
\bar{v}_t(x) =& \, b(x) + \int_E f(x \mid \xi) p_s(\xi) v_{t-1}(\xi) \, d\xi \label{eq:phd_prediction} \\
v_{t}(x) =& \, (1 - p_d(x \mid q)) \bar{v}_t(x) + \sum_{z \in Z_t} \frac{\psi_{z,q}(x) \bar{v}_t(x)}{\eta_z(\bar{v}_t)} \label{eq:phd_update} \\
\eta_z(v) =& \, c(z \mid q) + \int_E \psi_{z,q}(x) v(x) \, dx \label{eq:normalization} \\
\psi_{z,q}(x) =& \, g(z \mid x, q) p_d(x \mid q),
\end{align}
\label{eq:phd_filter}
\end{subequations}
where $\psi_{z,q}(x)$ is the probability of a sensor at $q$ receiving measurement $z$ from a target with state $x$ \edit{and \eqref{eq:normalization} is used within the prediction step \eqref{eq:phd_update}}.
\edit{The subscript $t$ refers to time.}
Note that $v(x)$ is \emph{not} a probability density function (PDF), but can converted into one by normalizing it with the expected number of targets in a sub-space of $E$,
\begin{equation}
p_t(x|E) = \frac{v(x)}{\int_E v(x)dx}.
\label{eq:p_t}
\end{equation}

Dames \cite{dames2020distributed} developed a distributed PHD filter in which each robot maintains the PHD within a unique subset, $\mathcal W_i$, of the environment while ensuring the distributed filtering scheme yields the same target estimation performance as its centralized counterpart.
Three algorithms then account for motion of the robots (to update the subsets $\mathcal W_i$), motion of the targets (in \eqref{eq:phd_prediction}), and measurement updates (in \eqref{eq:phd_update}).
In this paper, we adopt the same strategy for each robot to locally maintain its portion of the PHD.

\section{Optimized Space Partition}
\label{sec:partition}
We propose the novel \emph{normalized unused sensing capacity} in Section~\ref{subsec:unused}, which will be used to quantify the sensor heterogeneity in Section~\ref{sec:heterogeneity_level}.
After that, two space partitioning algorithms are developed based on power diagram (Section~\ref{subsec:power diagram}) and CCVD (Section~\ref{subsec:ccvd}).
Lastly, a control policy is proposed in Section~\ref{sec:poses} to optimize robot poses given the space partitioning schemes.

\subsection{Normalized Unused Sensing Capacity}
\label{subsec:unused}
We assume that each robot $s_i$ has a finite field of view (FoV) $F_i$, which could be different across the sensors $i$. 
Examples of $F_i$ include a wedge shape for a camera (\eg Figure~\ref{fig:footprint}) or a circle for a lidar.
Let $p_d(x|q_i,\theta_i)$ denote the probability of robot $s_i$ at position $q_i$ and with orientation $\theta_i$ detecting a target with state $x \in E$, which is the same model as in the PHD filter. 
The robot cannot detect targets outside its FoV $F_i$, \ie $p_d(x|q_i,\theta_i) = 0 \, \forall x \notin F_i$.
We assume that $p_d$ is time-invariant in the robot's local coordinate frame, but we make no other assumptions about the shape of $F_i$ or the functional form of $p_d$.
In practice, $p_d$ can be estimated using data-driven approaches \cite{dames2015experimental} and/or prior knowledge \edit{of} the sensor model.

We define the total detecting capability of robot $s_i$ as
\begin{equation}
D_i = \int_{F_i} p_d(x|q_i,\theta_i) \, dx,
\label{eq:det prob}
\end{equation}
which depends on the size of $F_i$ and the sensor's ability to detect information within $F_i$. 

In this paper, we consider sensor heterogeneity in terms of not only the sensing capability, such as FoV and detection accuracy, but also its current usage to track targets.
When a sensor detects a target and starts to track it, the sensor's sensing capability will be reduced.
We assume that we are tracking targets of finite size that cannot overlap, \edit{and that the target size is much smaller than the sensor's FoV}.
Therefore, \edit{for a robot $i$}, there exists some area $\mathcal{B}$ such that only 1 object can be in that area, \ie $\max \int_{\mathcal{B}} v(x) \, dx = 1$. 
Then 
\begin{align}
    &\max \int_\mathcal{B} \edit{p_d(x|q_i,\theta_i)} v(x) \, dx \nonumber \\
    &\leq \max \left(\max_{x \in \mathcal{B}} \edit{p_d(x|q_i,\theta_i)} \right) \int_\mathcal{B} v(x) \, dx \nonumber \\
    &= \max_{x \in \mathcal{B}} \edit{p_d(x|q_i,\theta_i)} \approxeq p_{d},
\end{align}
where the last approximate equality holds for small $\mathcal{B}$, such that $p_d$ is approximately constant over $\mathcal{B}$.
We can then write robot $s_i$'s FoV as the union of disjoint regions, $\mathcal{B}_k$, with the above property, so that $F_i = \cup_k \mathcal{B}_k$ and $\mathcal{B}_i \cap \mathcal{B}_j = \emptyset, \, \forall i \neq j$. 
Letting $p_{d,k} \approxeq \max_{x \in \mathcal{B}_k} p_d(x|q_i,\theta_i)$, we then define robot $s_i$'s maximum sensing capacity as
\begin{align}
    C_{\textrm{max},i}
        &= \mu \max \int_{F_i} p_d(x|q_i,\theta_i) v(x) \, dx \nonumber \\ 
        &= \mu \max \sum_{k} \int_{\mathcal{B}_k} p_d(x|q_i,\theta_i) v(x) \, dx \nonumber \\
        &\approxeq \mu \sum_{k} p_{d,k} 
        = \frac{\mu}{|\mathcal{B}|} \sum_{k} |\mathcal{B}| p_{d,k} \nonumber \\
        &= \frac{\mu}{|\mathcal{B}|} \int_{F_i} p_d(x|q_i,\theta_i) \, dx 
        = \frac{\mu \edit{D_i}}{|\mathcal{B}|}, \label{eq:det_max}
\end{align}
where \edit{$D_i$} comes from \eqref{eq:det prob} and $\mu$ is a tuning parameter associated with the maximum target density in the task space. 
For instance, as the expected maximum distance between the targets becomes larger, a smaller $\mu$ is selected to discount the maximum sensing capacity of a sensor.
Remark~\ref{remark:mu} discusses the selection of $\mu$ in more detail.

The expected number of target detections at $x$ is given by
\begin{equation}
p_{\textrm{exp}}(x) = \edit{p_d(x|q_i,\theta_i)p_t(x|E)},
\label{eq:p_exp}
\end{equation}
where $p_t(x)$ is defined in \eqref{eq:p_t}.
Thus, the total expected target detection \edit{probability} is given by
\begin{equation}
\begin{split}
C_{\textrm{exp},i} &= \int_{F_i} p_d(x|q_i,\theta_i) p_t(x|F_i) \,dx \\
& \overset{\eqref{eq:p_t}}{=} \frac{\int_{F_i}p_d(x|q_i,\theta_i)v(x) \, dx}{\int_{F_i}v(x) \, dx}.    
\end{split}
\label{eq:det_exp}
\end{equation}
\begin{definition}[Normalized Unused Sensing Capacity]
The relative \emph{normalized unused sensing capacity} with respect to the maximum target density for robot $s_i$, denoted $U_i$, is given by
\begin{align}
U_i &= C_{\textrm{max},i} - C_{\textrm{exp},i} \nonumber \\
    &= \int_{F_i}\left(\frac{\mu}{|\mathcal{B}|}-\frac{v(x)}{\int_{F_i}v(x)\, dx}\right) p_d(x|q_i,\theta_i) \, dx.
\label{eq:unused capacity}
\end{align}
\end{definition}
Note that \eqref{eq:unused capacity} can be easily modified by replacing $v(x)$ with a density function propagated via a different Bayesian filter.
The normalized unused sensing capacity quantifies current capacity for a sensor to track targets.
The larger the sensing capability of a robot, the higher the number of targets it can track. 
However, as the robot tracks an increasing number of targets, its \edit{remaining} sensing capability begins to decay.

\edit{\begin{remark}[Choice of $\mu$]
\label{remark:mu}
    The choice of $\mu$ is a free parameter. Setting it to a large value will make $C_{\rm max}$ large relative to $C_{\rm exp}$, resulting in all robots having similar unused capacities $U_i$. Picking $\mu = 0$ will result in robots only using the expected number of detections $C_{\rm exp}$ with no acknowledgement of total capacity. When the target density is unknown, we may choose $\mu = |\mathcal{B}|$ to account for the maximum possible number of target detection. Otherwise, $\mu$ can be set as the ratio of approximate $m$ and the number of $\mathcal{B} \in E$. 
\end{remark}}

\subsection{Power Diagram Implementation}
\label{subsec:power diagram}
To maximize the total detection probability of targets, we control the robots, considering their heterogeneity in spatial deployment.
To this end, we optimize both space partitioning and each sensor's location within its assigned partition.
Unlike the Voronoi diagram, which is suitable for the sensors with a homogeneous and isotropic sensing model, power diagrams \cite{aurenhammer1987power} are often used to compute the optimal dominance regions when the team has heterogeneous sensing models.

\begin{figure}[t]
\centering
\includegraphics[width=0.5\columnwidth]{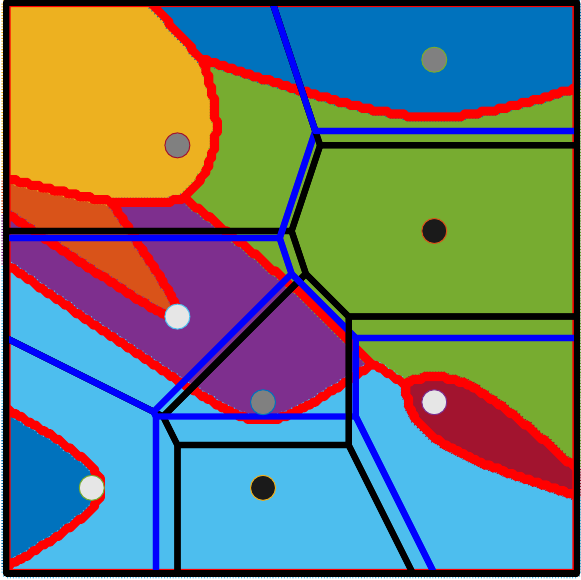}
\caption{Comparison of a Voronoi diagram (black lines), a power diagram (blue lines), and a CCVD (red curves and colored partitions). \edit{The darkness of each generation point (gray-scale dot) corresponds to its weight, i.e., power radius, with darker points having higher weights.} The three diagrams \edit{converge to the same solution} when the weights for all generation points are identical.}
\label{fig:diagrams}
\end{figure}

The power diagram is a variant of the standard Voronoi diagram that uses the power distance, 
\begin{equation}
    f(\|x - p_i \|) = \|x - p_i \|^2 - \rho(s_i)^2,
    \label{eq:power distance}
\end{equation}
where $\rho(s_i)$ is the weight or power radius of $s_i$, and $p_i$ is the generation point.  Figure~\ref{fig:diagrams} illustrates an example of the power diagram.
Existing power diagram-based approaches utilized sensor positions as the generation points, $p_i$, and the radii of the sensor FoVs as the weights, $\rho(s_i)$, to account for sensor heterogeneity \cite{arslan2016voronoi, pimenta2008sensing}. 
However, these approaches are limited to isotropic sensors.
On the other hand, our approach extends to heterogeneous anisotropic sensors.

We utilize the normalized unused sensing capacity $U_i$ to set the power radius in \eqref{eq:power distance}. 
This is a novel strategy to account for the heterogeneity in computing the dominance regions.
To achieve this, we proceed by expressing the optimization functional \eqref{eq:optimization} as
\begin{equation}
\mathcal{H}_p(Q, \mathcal{W}) = \sum_{i=1}^n \int_{\mathcal{W}_i} \left(\|x - p_i \|^2 - g(U_i)^2\right) \phi(x) \, dx,
\label{eq:h_p}
\end{equation}
where $g: \mathbb{R} \rightarrow \mathbb{R}$ is a mapping from the \edit{unused} sensing capacity to the power radius.
Since the normalized unused sensing capacity has units of area, we choose $g$ such that the resulting power radius, $g(U_i)$, is equal to the radius of a perfect (\ie $p_d = 1$) isotropic sensor with the same total sensing capability, $D_i$, \textit{i.e.}, $\pi g(U_i)^2 = U_i$.
Therefore we have
\begin{equation}
g(U_i) = \sqrt{\frac{U_i}{\pi}}.
\label{eq:g}
\end{equation}

Existing methods based on the power diagram, such as \cite{arslan2016voronoi, pimenta2008sensing, kwok2010deployment}, assume that the sensor's detection probability at $x$ in its assigned power partition is a non-increasing function of the distance from $x$ to the sensor.
In other words, the location of a sensor is the location that maximizes its detection probability of targets in its power partition.
Thus, it makes sense that they use the sensor location as the generation point, \ie let $p_i = q_i$.
However, this no longer holds true for anisotropic sensors.

Instead, we find the weighted centroid of the detection probability as
\begin{equation}
q_{\textrm{cod},i} = \frac{\int_{F_i} x p_d(x|q_i,\theta_i)\,dx}{\int_{F_i} p_d(x|q_i,\theta_i) \, dx},
\label{q_max}
\end{equation}
which we call the \emph{centroid of detection} (COD).
We use COD as the generation points for our power diagram, i.e., $p_i = q_{\textrm{cod},i}$.\footnote{Note that for an isotropic sensor the COD will be the same as the sensor position.}
Thus, the power partition of each robot becomes
\begin{equation}
\mathcal{W}_i 
= \{x \mid i = \argmin_{k = 1, \ldots, n} (\|x - q_{\textrm{cod}, k}\|^2 - g(U_k)^2)\}.
\label{eq:pv}
\end{equation}

\begin{remark}
For given $F_i$ and $p_d$ of a robot $s_i$, we can use \eqref{eq:det prob} to find an equivalent isotropic set $F_i'$ satisfying $\int_{F_i'} p_d(x|q_i,\theta_i) \, dx = D_i$.
This will map an arbitrary sensor model, characterized by $F_i$ and $p_d$, to a perfect isotropic sensor, characterized by a circular $F_i'$ with $p_d(x) = 1 ~\forall x \in F_i'$.
By choosing the appropriate mapping of the normalized unused sensing capacity $g(U_i)$, the weighted center of detection $q_{\textrm{cod},i}$ and the total sensing capacity $D_i$ are preserved as those of the original sensor model.
Hence, unused sensing capabilities of different sensors can be directly compared by their power radii, and so does the task spaces they should be assigned.
\end{remark}

\subsection{Capacity-Constraint Voronoi Diagram Implementation}
\label{subsec:ccvd}
The capacity-constraint Voronoi diagram (CCVD) \cite{michael2008capacity}, visualized in Figure~\ref{fig:diagrams}, computes the optimal task space assignment with the weight of each generation point as a hard constraint and yields closed-form optimal space partition in a discrete space.
This is done by two steps: initial cell assignment and cell swapping.
\emph{1) Firstly}, the task space is segmented into a finite set of regular grid cells $X = \{x_1, \cdots, x_{|X|} \}$, where each cell is identified by its center $x_i$.
The $i$th cell is indexed by $X[i]$. 
Then, we conduct an initial cell assignment satisfying a capacity constraint which specifies the maximum number of cells that can be assigned to each generation point.
\emph{2) Secondly}, we iteratively revise the cell assignment to minimize the total cost defined by
\begin{equation}
\sum_{x \in X} f(\|x - A(x) \|) = \sum_{x \in X} \|x - A(x) \|^2 - \sum_{x \in X} \rho(A(x))^2,
\label{eq:ccvd}
\end{equation} 
where $A(x)$ is the generation point assigned to $x \in E$, $f(\cdot)$ is a monotonically increasing function as introduced in \edit{\eqref{eq:optimization}}, and $\rho(\cdot)$ denotes the weight of a generation point as introduced in \eqref{eq:power distance}.
The right-most term in \eqref{eq:ccvd} is a constant for all assignments, and can therefore be omitted.
The cell assignment minimizing \eqref{eq:ccvd} results in a discrete power diagram where the number of cells in each power partition is equivalent to the capacity of its generation point.

\subsubsection{Initial Cell Assignment}
To implement the CCVD in the robot task assignment, $E$ is segmented into $|X|$ cells such that the size of each cell fits the maximum size of an individual target.
To take the sensor heterogeneity into consideration, we associate the \emph{capacity} of each robot $s_i$, \ie the number of cells assigned to $s_i$, denoted by $U_{\textrm{cap}, i}$, with its normalized unused sensing capacity $U_i$.
Therefore, we normalize $U_i$ to $U_{\textrm{cap}, i}$ by selecting a constant $I$ 
\begin{equation}
U_{\textrm{cap}, i} = I \cdot U_i, ~\forall i = 1,\ldots,n
\label{eq:const}
\end{equation}
at each discrete time step such that $\sum_{i = 1,\dots,n} U_{\textrm{cap}, i} = |X|$ and round $U_{\textrm{cap}, i}$ to an integer.
Similar to Section~\ref{subsec:power diagram}, we use $q_{\textrm{cod},i}$, \ie COD of each robot, as the CCVD generation points instead of robot's locations.
The initialization step requires all robots to synchronize $I$ in order to find $U_{\textrm{cap}, i}$ through \eqref{eq:const} in a distributed manner.
To achieve that, a distributed consensus protocol \cite{olfati2007consensus} is applied as outlined in Algorithm~\ref{alg:consensus}.
Initially, an equal number of cells is assigned to each robot $s_i$ to compute a temporary constant $\Tilde{I}_i$.
Then the robots reach a consensus on $\Tilde{I}_i = I$ via Equation~\ref{eq:consensus} to determine $U_{\textrm{cap}, i}$ using \eqref{eq:const} distributedly.
\begin{algorithm}[tbp]
\DontPrintSemicolon
\SetKwInOut{Input}{Input}
\SetKwInOut{Output}{Output}
\Input{$U_i, X$}
\Output{$\mathcal{W}^0_i$}
Initialize $U_{\textrm{cap}, i} \leftarrow |X|/n$ and set $\Tilde{I}_i \leftarrow U_{\textrm{cap}, i}/U_i$ \;
Find the neighbor set $\mathcal{N}_i$ \;
Update $\Tilde{I}_i$ using
\begin{equation}
\Tilde{I}_i \leftarrow \Tilde{I}_i + \sum_{j \in \mathcal{N}_i} (\Tilde{I}_j - \Tilde{I}_i) 
\label{eq:consensus}
\end{equation}
until $\Delta t$ time is up\;
Compute $U_{\textrm{cap}, i}$ using Equation~\ref{eq:const} \;
$\mathcal{W}^0_i \leftarrow \{X[\sum_{j=1}^{i-1} \cdot U_{\textrm{cap}, j}+1], ~\ldots, ~X[\sum_{j=1}^{i} \cdot U_{\textrm{cap}, j}]\}$
\caption{Distributed Initialization (Single Robot $s_i$ in One \edit{Iteration})} 
\label{alg:consensus}
\end{algorithm}


\subsubsection{Recursive Cell Assignment}
The original CCVD is constructed by iteratively swapping the cell assignment to all generation points, which is computationally expensive since each of the cells needs to be examined for the optimal assignment. 
In contrast, the enhanced approach by \cite{li2010fast} reduces the computational complexity without compromising the quality of the point distribution, by allowing a more efficient assignment strategy called median site swap, leading to faster convergence and lower time complexity. 
To illustrate, the median site swap method focuses on \edit{finding an optimal cell $\tau$ to as a distance reference to re-assign cells between the two robots $s_i$, and $s_j$}, without the need to examine every cell.

We propose a distributed cell assignment algorithm, outlined in Algorithm~\ref{alg:distributed}, based upon \cite[Algorithm 3]{li2010fast}.
Initially, an indicator $stable_i$ is set to false for a robot $s_i$, indicating that the robot has not been assigned the best set of cells that minimizes \eqref{eq:ccvd}.
Then the robot compares its ID with those of other robots from its neighbor set.
It \edit{undertakes the computation for} the neighbors with greater IDs by updating the assignment as outlined in Lines 4-18, \edit{while requesting updates to its capacity from neighbors with smaller IDs, as outlined} in Lines 19-22.

As an example, consider two robots $s_i,s_j$ with $i < j$. A \emph{serving} robot $s_i$ requests the capacity and the position from its served neighbor $s_j$ to conduct cell assignment, demonstrated as follows.
For this pair of neighboring robots, we aim to minimize the cost given by
\begin{equation}
\Delta e(x, q_{\textrm{cod},i}, q_{\textrm{cod},j}) := \|x-q_{\textrm{cod},i}\| - \|x-q_{\textrm{cod},j}\|.
\label{eq:key}
\end{equation}
In Lines 7-9, we calculate the cost defined in \eqref{eq:key} for robot $s_i$ to cover a cell and store it as a key in an array $P_{ij}$. 
Then, we need to find the optimal $U_{\textrm{cap}, j}$ cells out of the total $(U_{\textrm{cap}, i} + U_{\textrm{cap}, j})$ to assign for robot $s_j$ in order to minimize the cost, while assigning the rest to robot $s_i$. 
This is done in Line 10 by finding a cell $\tau$ which \edit{only allows} $U_{\textrm{cap}, j}$ cells to have $\Delta e(x_k, q_{\textrm{cod},i}, q_{\textrm{cod},j}) < \Delta e(\tau, q_{\textrm{cod},i}, q_{\textrm{cod},j})$. 
Physically, the cell $\tau$ is the median cell, \edit{\ie the cell with the median $\Delta e$ value,} of all cells assigned to both robots $s_i$ and $s_j$. 
After finding $\tau$, any cell that costs less than the cell $\tau$ is assigned to robot $s_j$; otherwise it is assigned to robot $s_i$ as outlined in Lines 11-15. 
Lines 16 and 17 is to keep track of the convergence of the assignment, which occurs when $\tau$ remains unchanged from the previous iteration.
On the other hand, the served robot sends its capacity and position to its serving neighbor to request its assignment $\mathcal{W}_i$, outlined in Lines 21-22.

One may notice that Algorithm~\ref{alg:distributed} yields uneven computational workload at each robot. 
In particular, robots with smaller IDs take on more computation tasks for the cell assignment than those with larger IDs.
In practice, one can assign smaller IDs to robots with higher computational resources.
\begin{algorithm*}[tbp]
\DontPrintSemicolon
\SetKwInOut{Input}{Input}
\SetKwInOut{Output}{Output}
\Input{$\mathcal{W}^0_i, X$}
\Output{$\mathcal{W}_i$}
$stable_i \leftarrow false$ \;
Find the neighbor set $\mathcal{N}_i$ \;
\For{\rm each robot $s_j \in \mathcal{N}_i$ with ID $j$}{
\If{$i < j$ }{ 
\While{$stable_i = false$}{
Request $\mathcal{W}^0_j, q_{\textrm{cod},j}$ from $s_j$ \edit{\Comment{Robot with smaller ID (serving robot) requests data from the other (served robot) and computes the partitions for both}} \; 
Initialize an array $P_{ij}$ \edit{\Comment{For storing cells and their costs, \ie keys}} \;
\For{\rm each cell $x_k \in \{\mathcal{W}^0_i,\mathcal{W}^0_j\} $}{
Insert $x_k$ into $P_{ij}$ with $\Delta e(x_k, q_{\textrm{cod},i}, q_{\textrm{cod},j})$ as its key \; 
}
Find $\tau \in P_{ij}$ so that only $U_{\textrm{cap}, j}$ cells have $\Delta e(x_k, q_{\textrm{cod},i}, q_{\textrm{cod},j) } < \Delta e(\tau, q_{\textrm{cod},i}, q_{\textrm{cod},j})$ \;
\For{\rm each cell $x_k \in P_{ij} $}{
\If{\rm $\Delta e(x_k, q_{\textrm{cod},i}, q_{\textrm{cod},j}) < \Delta e(\tau, q_{\textrm{cod},i}, q_{\textrm{cod},j})$}{
Assign $x_k$ to $\mathcal{W}_j$ \edit{\Comment{Assign the cell to $s_j$ if it costs less than the median cell}}
}
\Else{Assign $x_k$ to $\mathcal{W}_i$ \edit{\Comment{Otherwise, assign it to $s_i$}}}
}
\edit{\If{\rm $\tau$ unchanged from last iteration}{
$stable_i \leftarrow true$ \edit{\Comment{Reach the steady state and terminate}}
}}
}
$\mathcal{W}_i \leftarrow \textrm{Swapped} ~\mathcal{W}^0_i$ \edit{\Comment{Update its own partition}} \;  
Send $\mathcal{W}_j \leftarrow \textrm{Swapped} ~\mathcal{W}^0_j$ to $s_j$  \edit{\Comment{Send updated partition to the other robot}} \; \;
}
\Else{
Send $\mathcal{W}^0_i, q_{\textrm{max},i}$ to $s_j$ \edit{\Comment{Robot with larger ID sends data to the other and wait for the partition}} \; 
Request $\mathcal{W}_i$ from $s_j$ \;
}
}
\caption{Distributed Cell Swapping (Single Robot $s_i$ in One \edit{Iteration})} 
\label{alg:distributed}
\end{algorithm*}

\subsection{Optimized poses}
\label{sec:poses}
\begin{algorithm}[tbp]
\DontPrintSemicolon
\While{active}{
Find optimized space partition $\mathcal{W}_i$ \;
Update $\phi(x) = v(x)$ for $x \in F_i$ using \eqref{eq:phd_filter}\;
Send $\phi(x), x \in F_j$ to neighbors for all $\{s_j~|~F_i \cup \mathcal{W}_j \neq \emptyset\}$ \;
Receive $\phi(x), x \in \mathcal{W}_i$ from neighbors for all $\{s_j~|~F_j \cup\mathcal{W}_i \neq \emptyset\}$ and compute $C_{\mathcal{W}_i}$ via \eqref{eq:mass}\;
Execute control $u_i$ and \edit{$\omega_i$} via \eqref{eq:u_i} \;}
\caption{Distributed Control (Single Robot $s_i$)} 
\label{alg:control}
\end{algorithm}
Once the optimized partition is retrieved, each robot must move to its optimized pose to improve the detection probability.
By adopting the same analytical approach presented in \cite{cortes2004coverage}, we can compute the partial derivative of $\mathcal{H}_p(Q, \mathcal{W})$ with respect to $q_{\textrm{cod},i}$ as
\begin{equation}
\frac{\partial\mathcal{H}_p(Q)}{\partial q_{\textrm{cod},i}} = 2M_{\mathcal{W}_i}(q_{\textrm{cod},i} - C_{\mathcal{W}_i}),
\label{eq:partial}
\end{equation}
where $M_{\mathcal{W}_i}$ and $C_{\mathcal{W}_i}$ are \edit{the mass and center of mass associated with the region $\mathcal{W}_i$, respectively}, defined as
\begin{equation}
\begin{aligned}
M_{\mathcal{W}_i} =& \int_{\mathcal{W}_i} \phi(x) \, dx, \\
C_{\mathcal{W}_i} =& \frac{1}{M_{\mathcal{W}_i}}\int_{\mathcal{W}_i} x\phi(x) \, dx.
\label{eq:mass}
\end{aligned}
\end{equation}
Thus, as \edit{$q_{\textrm{cod},i}$ is recursively driven} to $C_{\mathcal{W}_i}$, the partial derivative approaches to $0$ and thus sensing capability of $s_i$ in $\mathcal{W}_i$ is optimized.
The control inputs in \eqref{eq:dynamic} for $s_i$ \edit{are} then given by
\begin{equation}
\begin{aligned}
u_i =& \, \|C_{\mathcal{W}_i}-q_{\textrm{cod},i}\| (dt)^{-1},\\
\Delta\theta =& \, \textrm{ang}(C_{\mathcal{W}_i}-q_{\textrm{cod},i}) - \theta_i,\\
\edit{\omega_i} =& |\Delta\theta| (dt)^{-1} \textrm{sgn}(\Delta\theta-\theta_i) , 
\label{eq:u_i}
\end{aligned}   
\end{equation}
where $\textrm{ang}(\cdot)$ denotes the angle of a position vector in global frame.

Algorithm~\ref{alg:control} outlines the distributed control algorithm for each robot.
At each time step, the robot first finds the optimized partition using one of the approaches explained in Sections~\ref{subsec:power diagram} and~\ref{subsec:ccvd}, and maintains the PHD within its partition using the same strategy as in \cite{dames2020distributed}, which requires the exchange of local PHD with its neighbors.
Then, the robot computes its control using \eqref{eq:u_i}.

\section{Performance Analysis}
\label{sec:performance}
\subsection{Computational Complexity}
Our distributed algorithms for the space partitioning using a power diagram, as described in Sections~\ref{subsec:power diagram} and~\ref{subsec:ccvd}, are similar to distributed computation methods for constructing a Voronoi diagram. Resorting to the computational complexity analysis reported in \cite{cao2003distributed}, the complexity of such methods is $\bigO{\left(\log{ |\mathcal{N}_i|}\right)}$ and, hence, so \edit{is} ours, where $|\mathcal{N}_i|$ is the number of neighbors of each robot $s_i$. 

In the CCVD implementation described in Section ~\ref{subsec:ccvd}, Algorithm~\ref{alg:distributed} requires each of the $n$ robots to find the optimal assignment of the total $|X|$ cells.
For the worst-case complexity analysis, we assume that each robot is a neighbor of all the other robots, resulting in $n - 1$ execution of the loop in Lines 3-21.
The most time-consuming part of the algorithm is in finding the cell $\tau$, \ie the median point in the unordered array $P_{ij}$. 
In our implementation, we employ Hoare's quick-select algorithm \cite{Hoare65} to find the median value $\tau$ by choosing a pivot randomly and dividing the array into two parts, \ie one with values less than or equal to the pivot and the other with the values greater than the pivot.  
The median value is either included in the former part if this array has $(n+1)/2$ elements or more, \edit{where} $n$ is the number of elements in the array, or the latter part otherwise.
This divide-and-conquer algorithm allows the algorithm to run in linear time on average, with the time complexity of $\bigO{\left(\frac{|X|}{n}\right)}$ \cite{li2010fast}. 
Hence, the total time complexity for each single neighbor iteration is $\bigO{\left(\frac{|X|}{n}\right)}$. 
Considering that the robot has $n - 1$ neighbors by assumption, the complexity for a single robot iteration is $\bigO{\left(|X|\right)}$.

Therefore, the distributed construction of power diagrams has lower computational complexity compared to the construction of CCVDs. 
Additionally, the complexity of constructing power diagrams \edit{depends} only on the number of neighbors, making it a better choice when the computational power of the robots is \edit{limited}, despite the fact that it provides only a near-optimal solution for space partitioning.
One may choose to use the most suitable algorithm depending on the available computational resources and trade-off between computational time and tracking accuracy. 

\subsection{Communication Load}
We assume that the communication between each robot and its neighbors is lossless and delay-free.
The bandwidth requirement for the robot network is mainly determined by the number of neighbors each robot has and both the size and the frequency of data exchange. 
On the one hand, given a fixed data size and communication frequency, the worst-case bandwidth requirement occur when all robots are neighbors of one another \edit{, in which case} $n \left(n-1\right)$ non-overlapping communication links are required, 
where $n$ is the number of robots. 
In fact, this worst case assumption is unlikely to happen as the robots are expected to spread out over a large task space. 

On the other hand, given a fixed number of communication channels, each robot only needs to exchange information about its space assignment (either power partition vertices or CCVD cell sets), location, and local PHD to its neighbors.
The Voronoi methods, at least with constant weighting function, typically result in each robot having 6 neighbors in a hexagonal packing structure \cite{yan2011computing}.
Hence, the bandwidth requirement is low, and will not increase as the environmental dimension (number of robots and targets, size of the task space, etc) increases.

\edit{To analyze the communication load of the space partitioning algorithms, we consider the data size of the information exchanged between robots. For the power diagram, the data size is determined by the number of vertices in the power partition, which is proportional to the number of cells in the task space. Each vertex is represented by a 2D position vector, which requires 8 bytes per vertex for a double-precision floating point number. For the CCVD, each robot shares its current cell assignments, capacity, and local information about detected targets. The data size is proportional to the number of cells in the task space and the number of detected targets. Each cell is represented by a 2D position vector, which requires 8 bytes per cell. Each detected target is represented by a 2D position vector and a detection probability, which requires 12 bytes per target. Hence, for an average of 6 neighbors, 10 targets per robot, and a frequency of 1 Hz, the bandwidth requirement for the power diagram is $8 \times 6 \times 1 = 48$ bytes per second, and for the CCVD is  $\left(8  + 12 \times 10\right) \times 6 \times 1 = 768$ bytes per second.}

\edit{Currently, robots share all information with their neighbors in every iteration. However, the communication frequency and bandwidth requirements can be further reduced through selective sharing approaches, where re-partitioning only happens when robot positions change significantly.}

\subsection{Performance Metrics}
\subsubsection{Tracking Accuracy} To assess the tracking performance, we use the first order Optimal SubPattern Assignment (OSPA) metric \cite{schuhmacher2008consistent}, which is a widely-adopted metric to evaluate the performance of MTT approaches.
Given two sets $X$ and $Y$ (representing the true and estimated target locations), the tracking error is defined as\footnote{Without loss of generality, we assume that $|X| = m \leq |Y| = n$ holds. \edit{In other words, $X$ represents either the true or estimated target set, whichever is smaller.}}
\begin{multline}
d(X,Y) = \\ \left( \frac{1}{n} \min_{\pi \in \Pi_{n}} \left( \sum_{i = 1}^{m} d_c(x_i, y_{\pi(i)})^p + c^p (n - m) \right) \right)^{1/p}.
\label{eq:ospa}
\end{multline}
The constant $c$ is a cutoff distance, $d_c(x,y) = \min(c, \|x - y\|)$, and $\Pi_n$ is the collection of all permutations of the set $\{1, 2, \ldots, n\}$.
The larger the value of $p$ is, the more the outliers are penalized.
Eq. \eqref{eq:ospa} computes the average matching error between true and estimated target locations considering all possible assignments between elements $x \in X$ and $y \in Y$ that are within distance $c$.
This can be efficiently computed in polynomial time using the Hungarian algorithm \cite{kuhn1955hungarian}.
Note that the \emph{lower} the OSPA value, the more accurate the tracking of the targets.

\subsubsection{Heterogeneity Level} 
\label{sec:heterogeneity_level}
We define a measure of the heterogeneity in a team as follows.
\begin{definition}[Heterogeneity Level]
The \emph{heterogeneity level} of a sensing network $S = \{ s_1, s_2, ..., s_n \}$, denoted by $L(S)$, is given by the standard deviation of the power radius of each sensor in $S$, \ie
\begin{equation}
L(S) = \sqrt{\frac{1}{n}\sum_{i=1}^n(g(C_{max,i}) - \bar{g})^2}
\label{eq:level}
\end{equation}
where $\bar{g} = \frac{1}{n}\sum_{i=1}^n g(C_{max,i})$, where the function $g$ is given in \eqref{eq:g} and $C_{max, i}$ is the maximum sensing capacity of robot $s_i$.
\end{definition}

\begin{definition}[Total Sensing Capacity]
The \emph{total sensing capacity}, of a sensing network $S$ is given by
\begin{equation}
C(S) = \sum_{i=1}^n C_{\textrm{max},i}.
\label{eq:total}
\end{equation}
\end{definition}

\section{Simulation Results}
\label{sec:results}
\begin{figure}[tbp]
\centering
\subfloat[Type 1 shape]{
	\includegraphics[width=0.3\columnwidth]{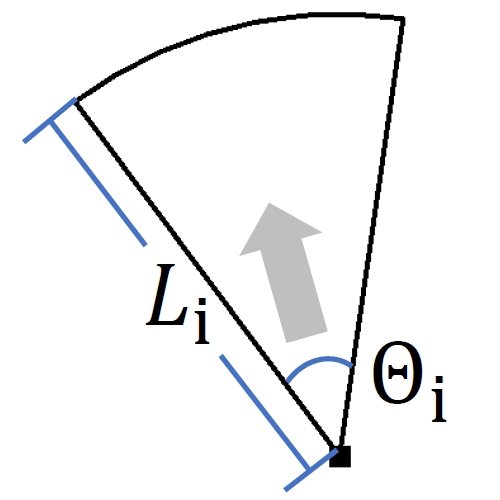}
   \label{fig:footprint1}}
   \hspace{1.5cm}
\subfloat[Type 2 shape]{
    \includegraphics[width=0.3\columnwidth]{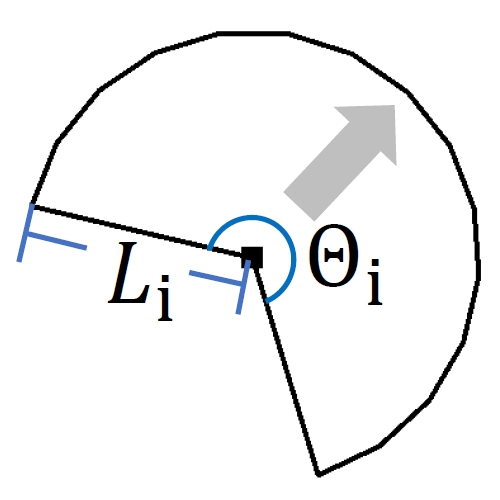}
   \label{fig:footprint2}
}
\caption{Two types of sensors used in the simulations. Type 1 and type 2 have viewing angles of $45^{\circ}$ and $240^{\circ}$, respectively. Black squares represent the location of sensors. The viewing angles, radii, and forward directions of both FoVs are indicated in the figures.}
\label{fig:footprint}
\end{figure}

We conduct simulations using both ROS and \textsc{MATLAB} to validate our proposed multi-robot multi-target tracking framework. 
For concise presentation, all simulations adopt sensors with wedge-shaped FoVs whose shape is parameterised by a viewing angle $\Gamma_i$ in the forward direction, which is the same as the orientation of a robot, and a radius $L_i$. 
Cameras and lidars have such model.
Two examples are visualized in Figure~\ref{fig:footprint}.
The probability of target detection for these FoVs is defined by
\begin{equation}
    p_d(x|q_i, \theta_i) = \begin{cases}
        f_{d,i}(\Delta L) & \text{if } x \in F_i \\
        0 & \textrm{otherwise},
    \end{cases}
\end{equation}
where $\Delta L = \|x-q_i\|$, and $f_{d,i}(\Delta L)$ is the probability density function that assigns the target detection probability given the distance $\Delta L$ between the target and robot $s_i$. Specific implementations of $f_{d,i}$ used in simulations are provided in Table~\ref{table:tb3_sensors}.

Unlike the most existing work such as \cite{pimenta2008sensing, arslan2016voronoi, kwok2010deployment, laventall2008coverage, stergiopoulos2012autonomous, kantaros2016distributed},
in this paper, we estimate the target distribution $\phi(x)$ online using the PHD filter. 
Hence, our problem is significantly more challenging since the robots need to perceive and track targets simultaneously.
We will compare our new approach against the standard Voronoi based method and the power diagram based algorithm proposed in our previous conference paper \cite{chen2021distributed} to provide a baseline for performance comparison.

\begin{table}[tbp]
\centering
\caption{TurtleBot3 Sensors}
\label{table:tb3_sensors}
\begin{tabular}{| c || c | c | c | c | c | c |}
\hline
\backslashbox{Types}{Specs} & \thead{$\Theta_i$\\ (deg)} & \thead{$L_i$ \\(\unit{m}) } & $f_{d,i}(\Delta L)$ & $C_{max}$\\
\hline
\hline
\textrm{1} & 270 & 3 & $0.99 - 0.1\cdot\Delta L$ & 1.675\\
\hline
\textrm{2} & 360 & 3 & $0.99 - 0.067\cdot\Delta L$ & 2.422\\ 
\hline
\textrm{3} & 90 & 3 & 0.99 & 0.700\\
\hline
\textrm{4} & 90 & 2.5 & $0.99 - 0.1\cdot\Delta L$ & 0.404\\
\hline
\textrm{5} & 360 & 2 & 0.99 & 1.257\\
\hline
\end{tabular}
\end{table}

\subsection{Qualitative Results}
\label{subsec:qualitative}
\begin{figure}[tbp]
\centering
\includegraphics[width=0.9\columnwidth]{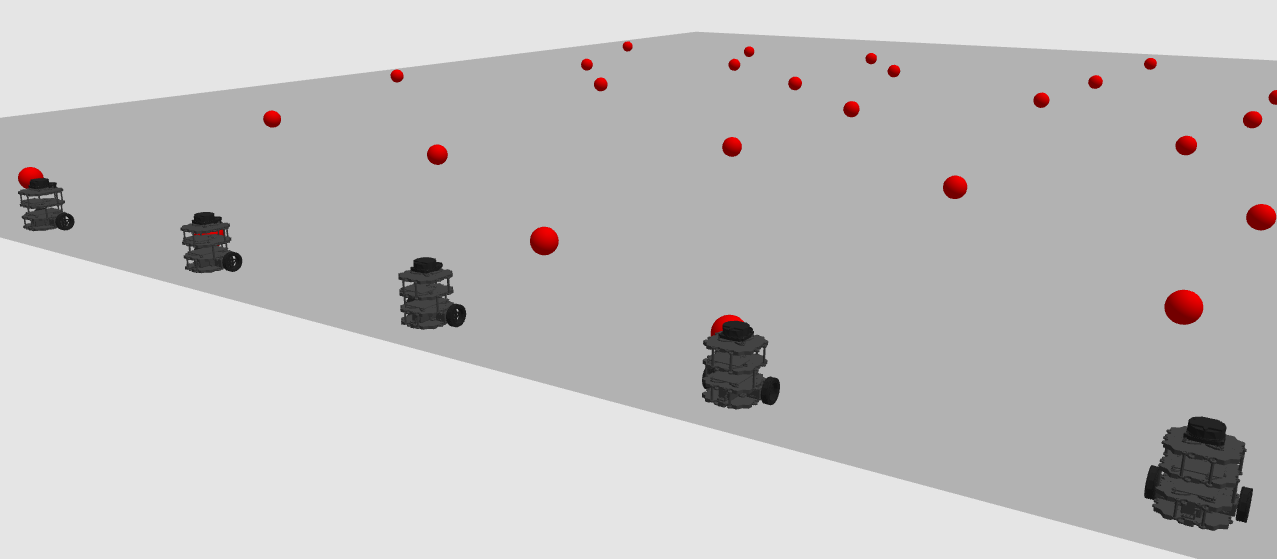}
\caption{Visualization of a simulation environment in Gazebo with 5 TurtleBot3 Burger wheeled robots and multiple moving targets, represented with red balls.}
\label{fig:gazebo_env}
\end{figure}

First, we \edit{apply CCVD implementation and} show the result from a single run using 5 TurtleBot3 Burger differential drive robots, namely $s_1, \ldots, s_5$, tracking 40 holonomic moving targets in a $\unit[10]{m} \times \unit[10]{m}$ obstacle-free \edit{square} task space as visualized in Figure~\ref{fig:gazebo_env}. 
The simulation was implemented using Ubuntu 18.04 with ROS Melodic. 
The robots move at a maximum linear velocity of $\unit[0.2]{m/s}$ and a maximum angular velocity of $\unit[1]{rad/s}$, and they are able to localize themselves using position data retrieved from ROS topic \texttt{/odom} \edit{, \ie the ground truth state from Gazebo}.
Our MR-MTT algorithm recursively generates next waypoints for the robots, and they navigate through the waypoints using the \texttt{move\_base} ROS package which uses the dynamic window approach (DWA) as a local planner and Dijkstra’s algorithm as a global planner \cite{zheng2021ros}.

All robots are equipped with heterogeneous range-bearing sensors with specifications described in Table~\ref{table:tb3_sensors}.
The standard deviation of range and bearing measurements for all sensors is \unit[0.04]{m} and $0.1^\circ$, respectively.
Since target density is low relative to the size of $E$ and the FoV of robots, we select $\mu = 0.1$ in \eqref{eq:det_max} for all robots.
The task space is discretized into $50\times50$ cell for both PHD representation and task space assignment. 
We assume that only one target can occupy each $\unit[{0.2}]{m} \times \unit[{0.2}]{m}$ cell.

As explained in Section~\ref{sec:intro}, an important application of the MR-MTT is in tracking targets to study collective behaviours of animal groups and crowd dynamics of pedestrians.
Motivated by related works in modeling their collective motion \cite{reynolds1987flocks}, we apply Boids algorithm, which has been used to study \edit{the emergent flocking behaviors arising in social animal groups from combinations of separation, alignment, and cohesion behaviors}, to simulate the moving targets. 

In our simulations, a target may exit the task space in which case another target will immediately enter into the space to maintain a constant number of targets.
Targets move at a maximum speed of $\unit[0.2]{m/s}$.

\begin{figure*}[tbp]
\centering
\subfloat[Gazebo 0-2'30'']{
	\includegraphics[width=0.53\columnwidth]{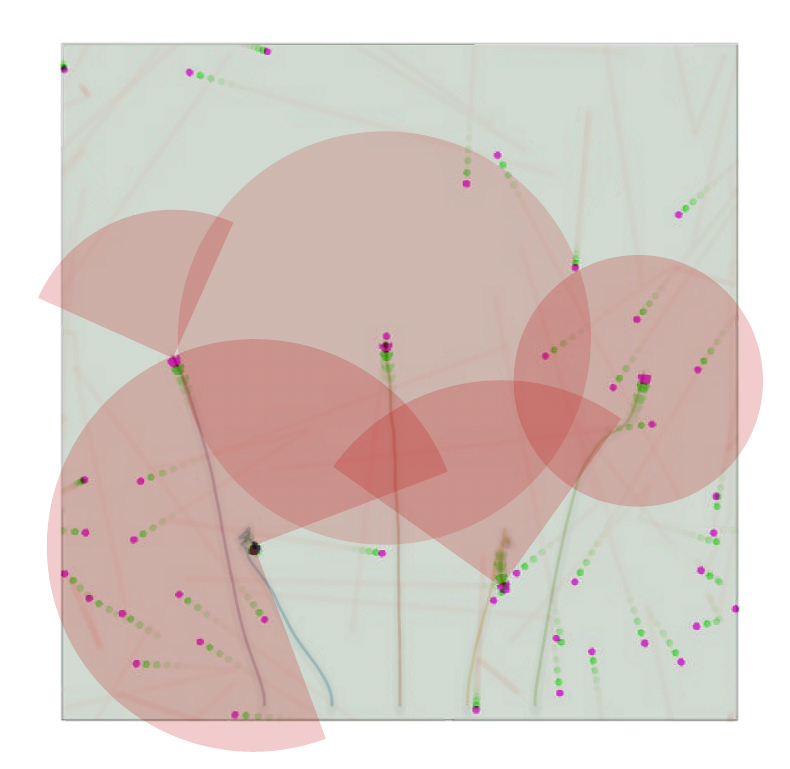}
	\label{fig:0_230}
} \hspace{0.16cm}
\subfloat[RViz 0'']{
	\includegraphics[width=0.5\columnwidth]{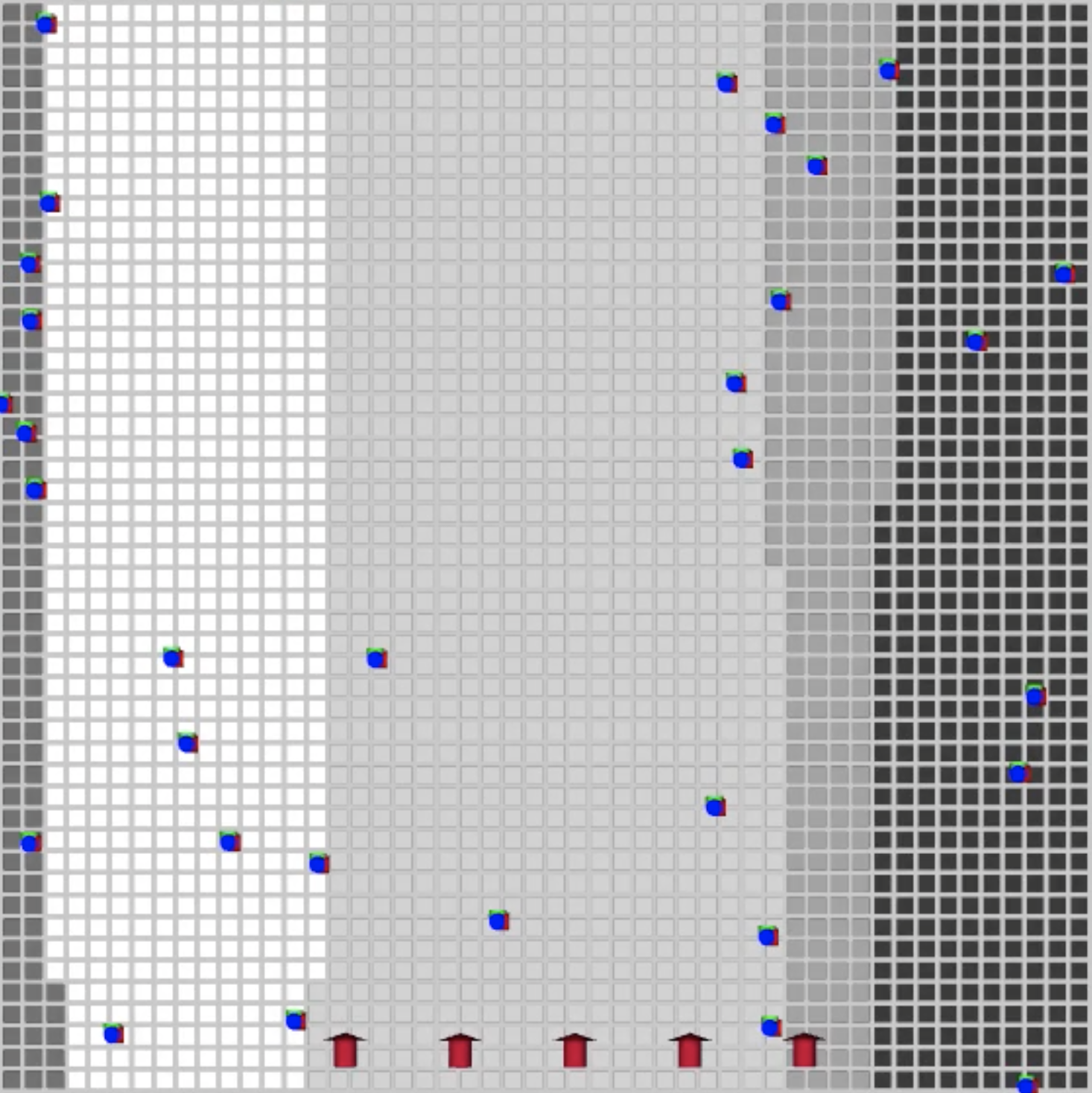}
	\label{fig:rviz_0}       
}  \hspace{0.4cm}
\subfloat[RViz 2'30'']{
	\includegraphics[width=0.5\columnwidth]{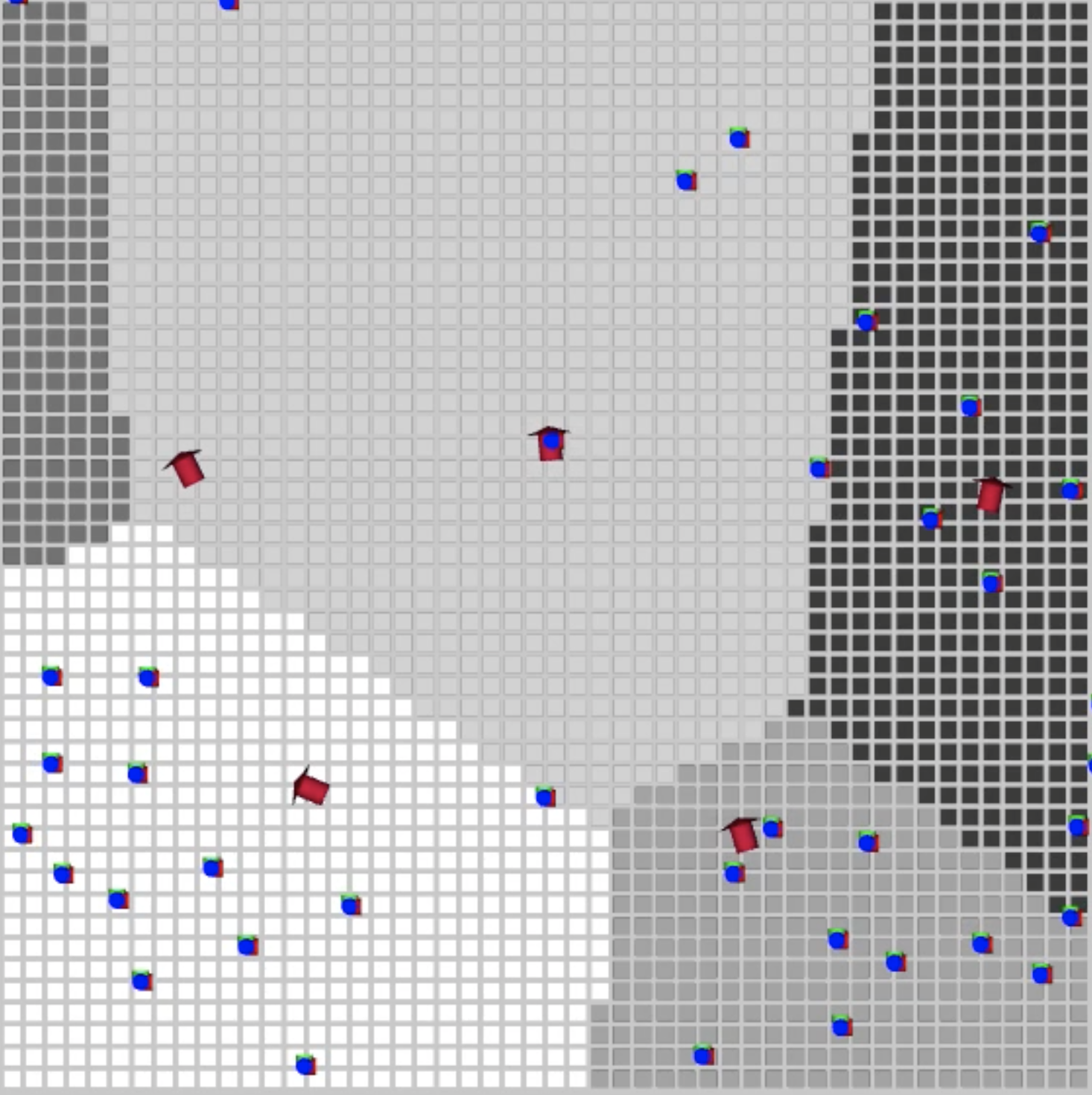}
	\label{fig:rviz_230}
} \\
\subfloat[Gazebo 3'40''-4'06'']{
	\includegraphics[width=0.51\columnwidth]{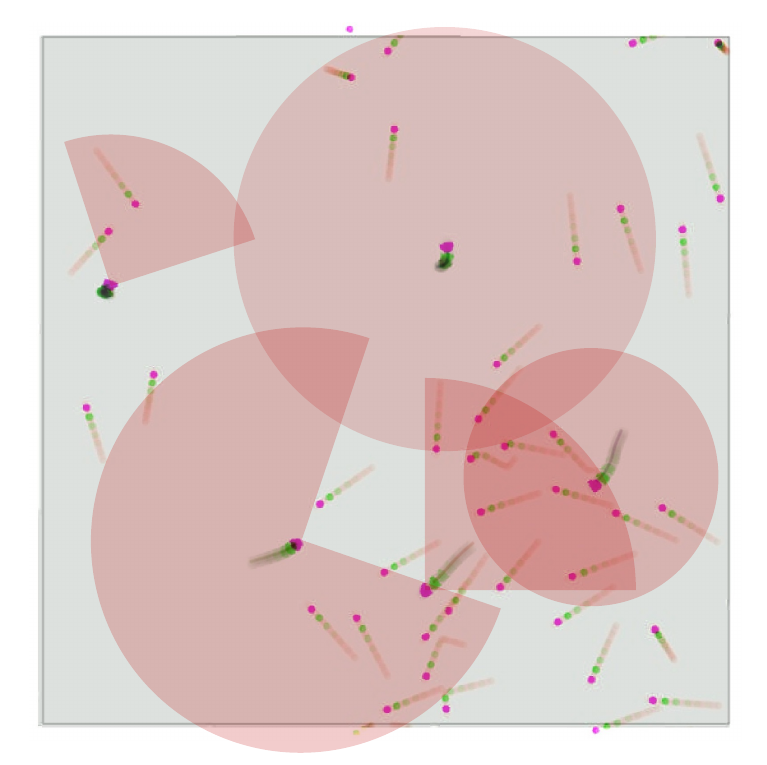}
	\label{fig:340_406}
} \hspace{0.3cm}
\subfloat[RViz 3'40'']{
	\includegraphics[width=0.5\columnwidth]{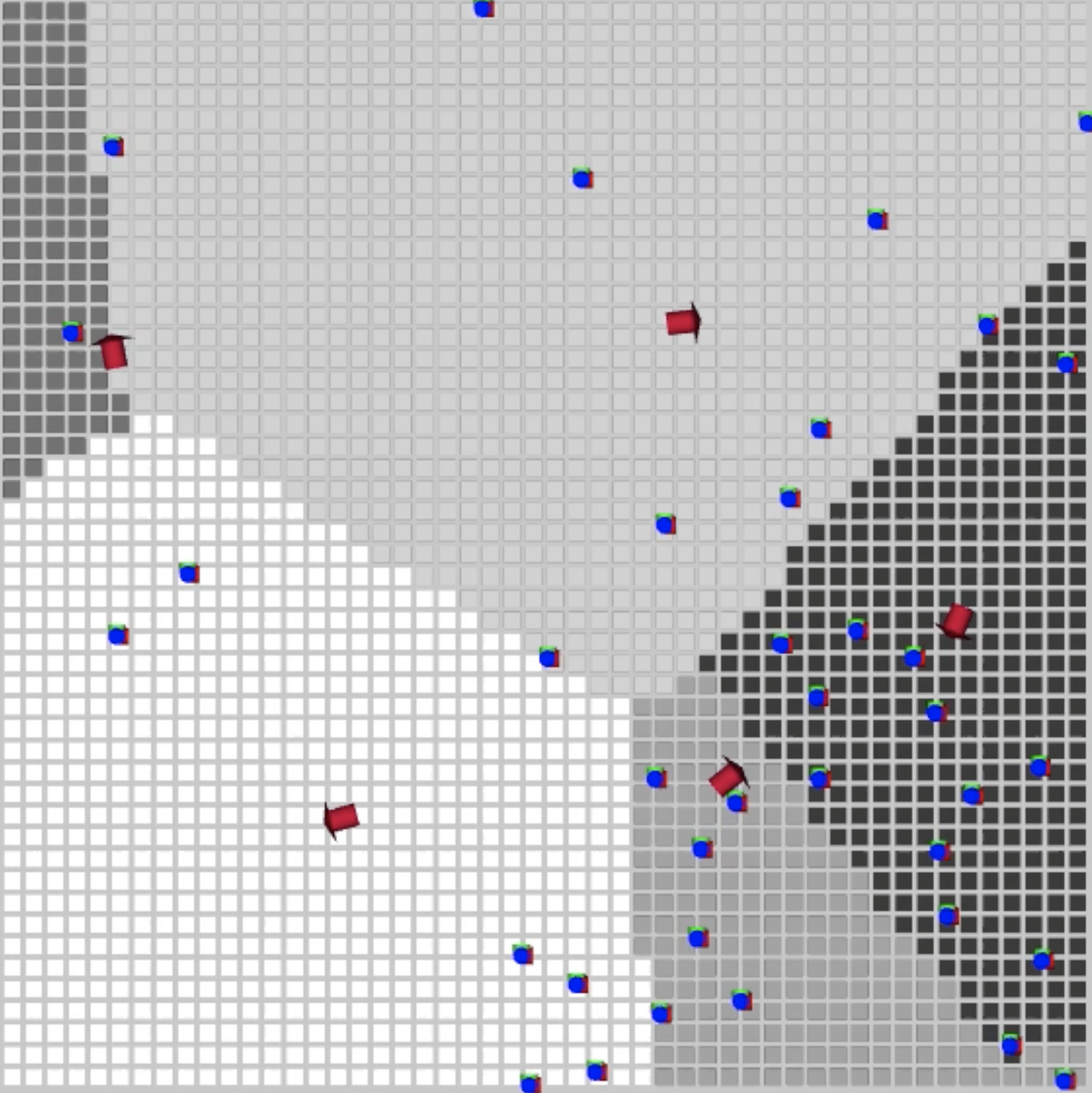}
	\label{fig:rviz_340}
}  \hspace{0.4cm}
\subfloat[RViz 4'06'']{
	\includegraphics[width=0.5\columnwidth]{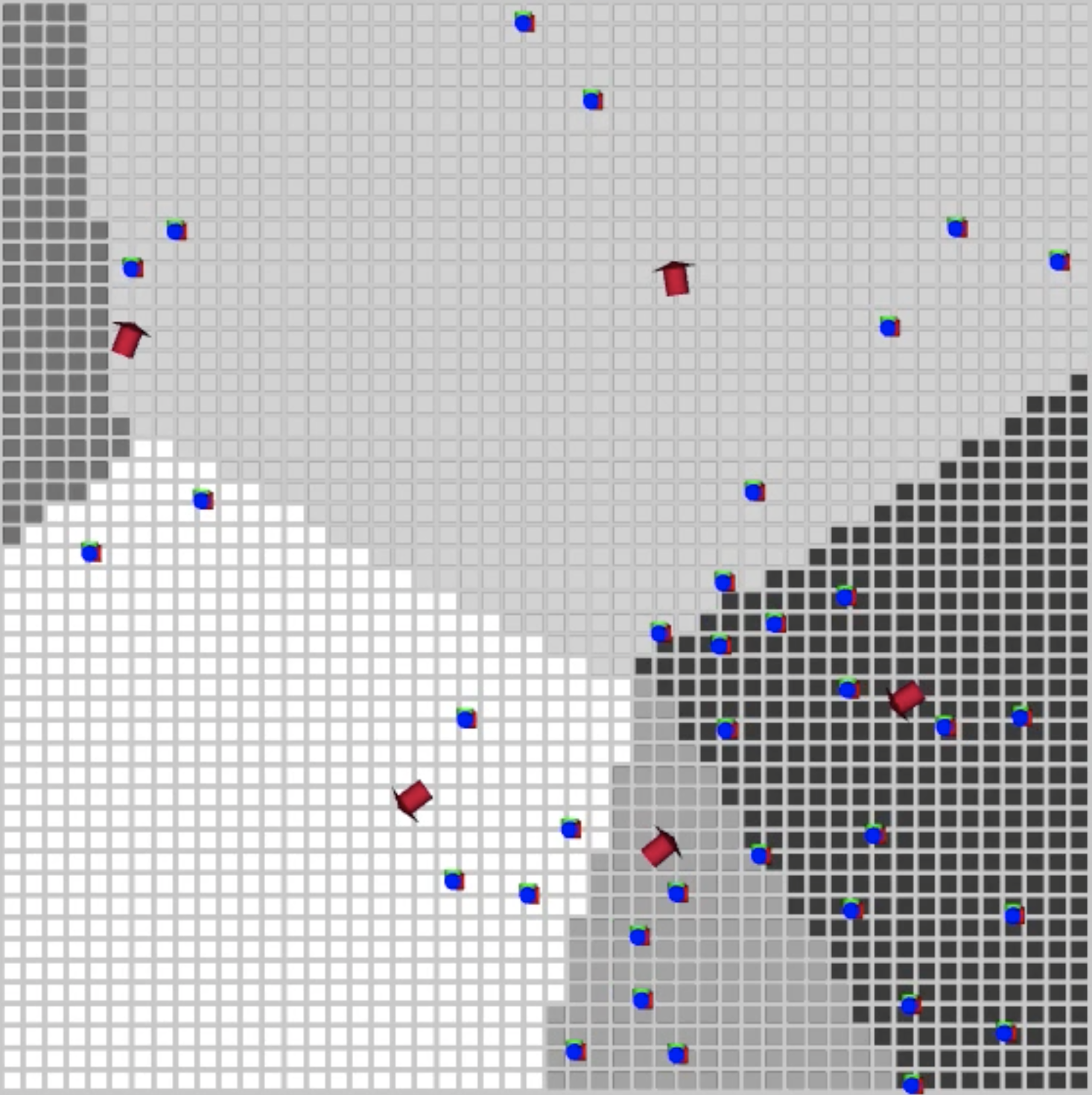}
	\label{fig:rviz_406}
} \\
\subfloat[Gazebo 4'30''-5'12'']{
	\includegraphics[width=0.58\columnwidth]{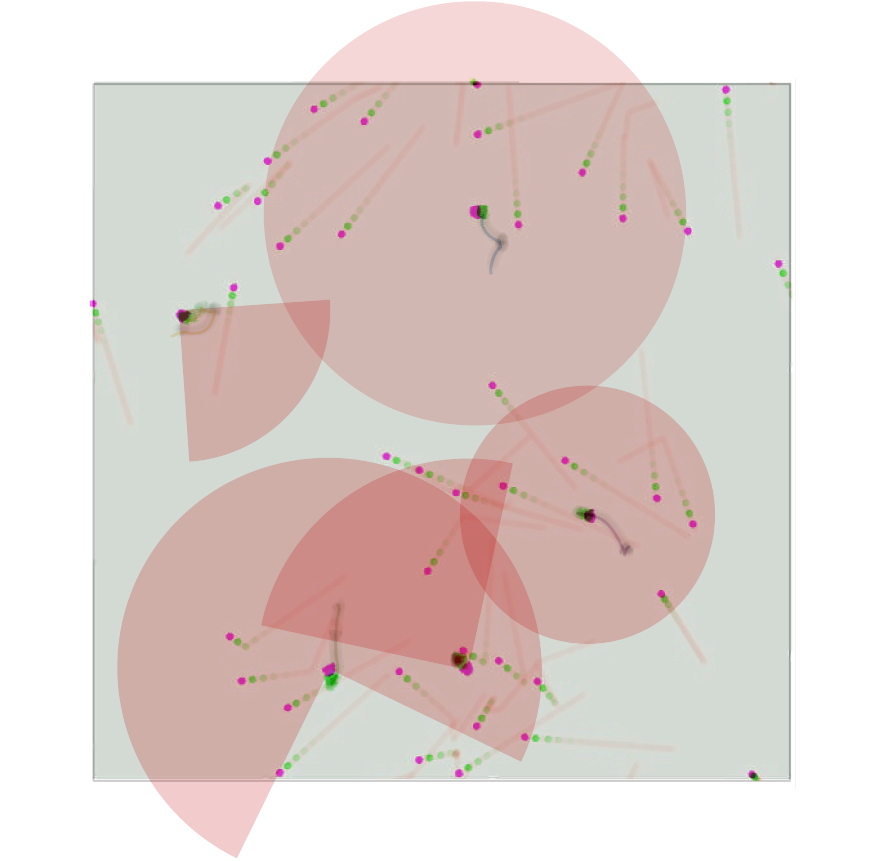}
	\label{fig:430_512}
}
\subfloat[RViz 4'30'']{
	\includegraphics[width=0.5\columnwidth]{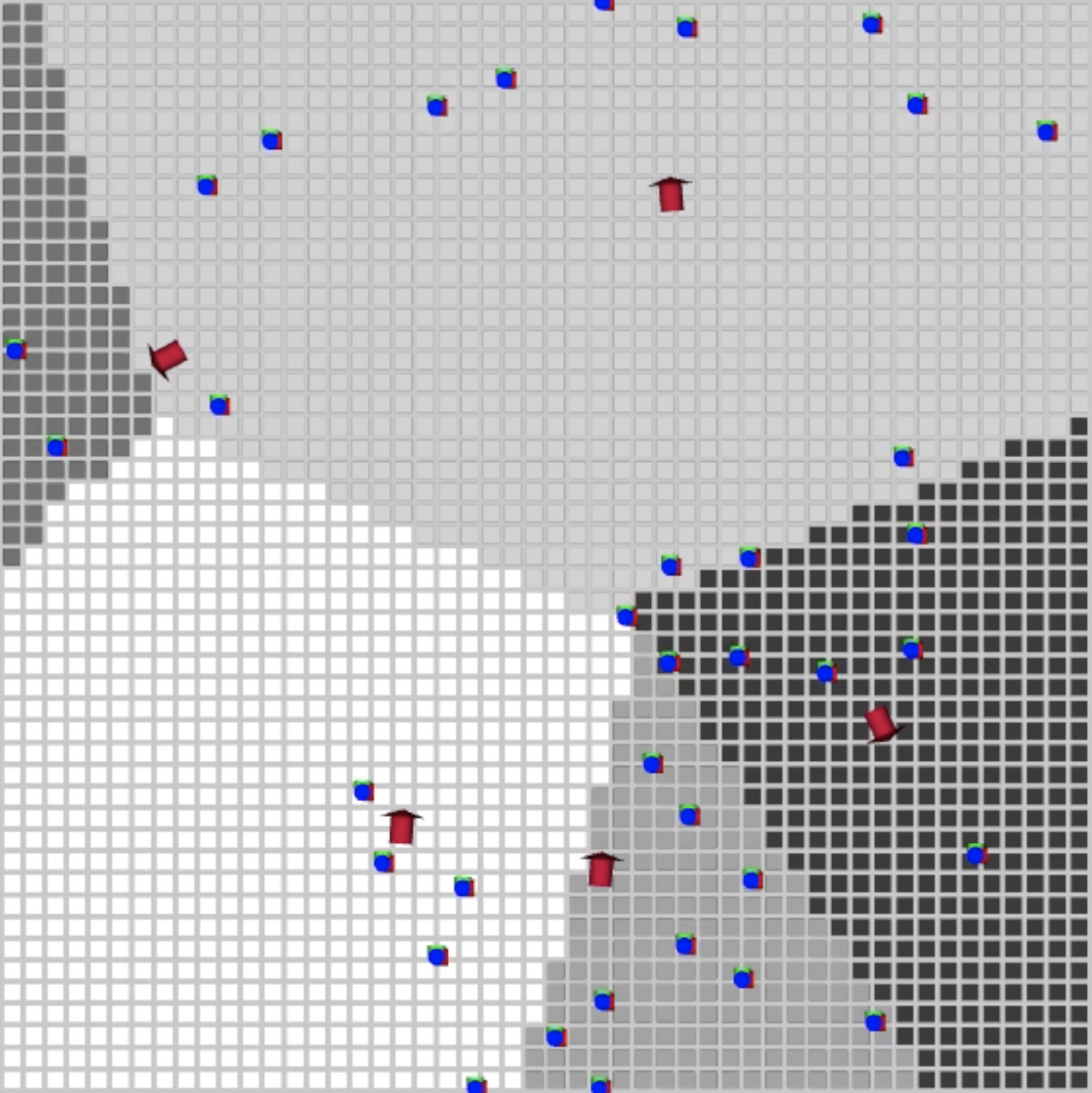}
	\label{fig:rviz_430}
}  \hspace{0.4cm}
\subfloat[RViz 5'12'']{
	\includegraphics[width=0.5\columnwidth]{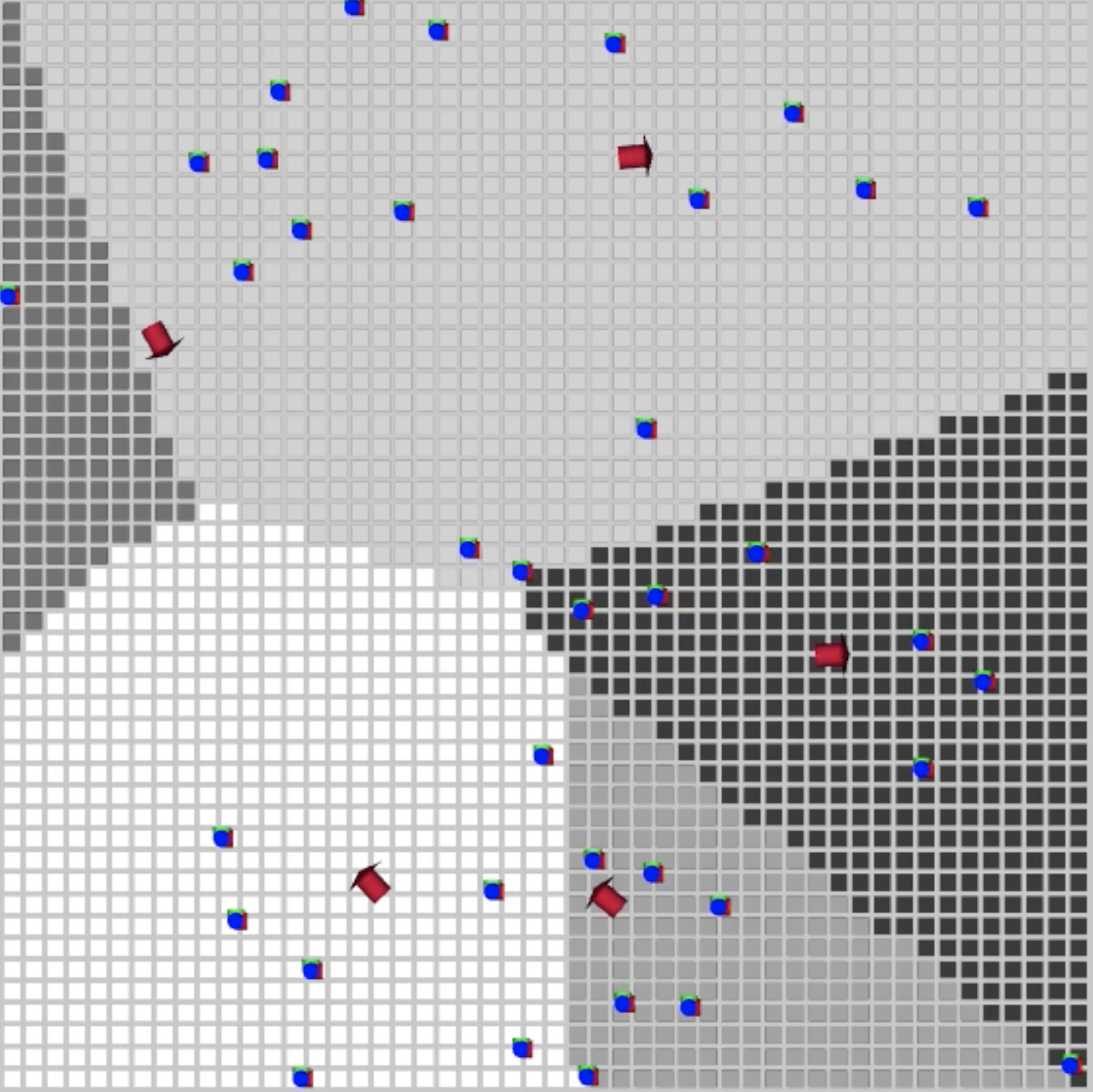}
	\label{fig:rviz_512}
} \\
\subfloat[Gazebo 5'20''-6'30'']{
	\includegraphics[width=0.53\columnwidth]{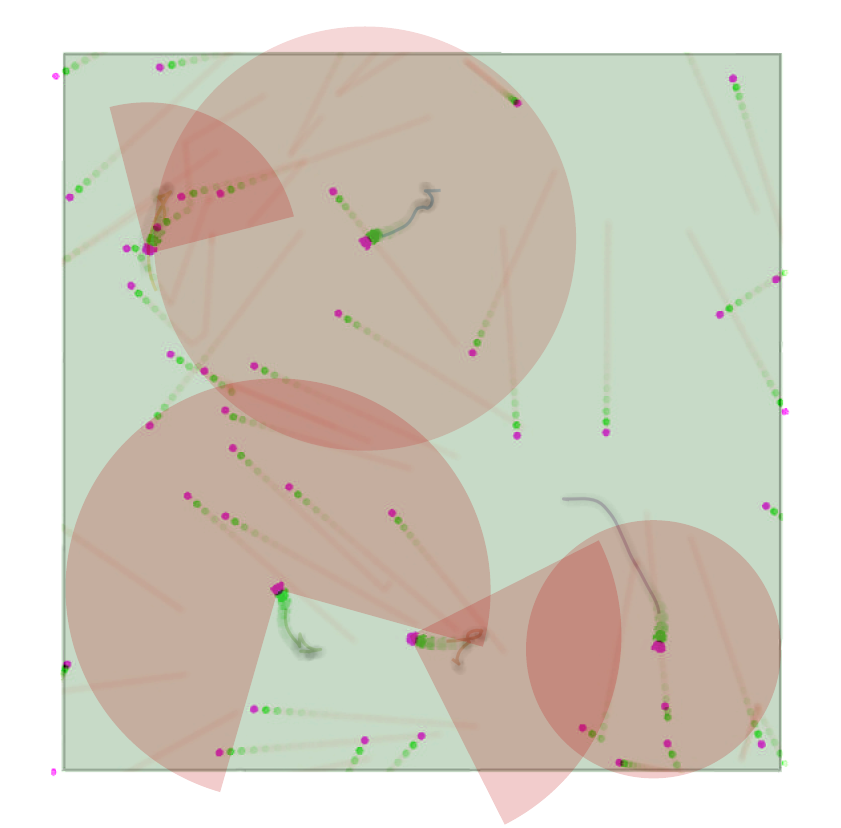}
	\label{fig:520_630}
} \hspace{0.2cm}
\subfloat[RViz 5'20'']{
	\includegraphics[width=0.5\columnwidth]{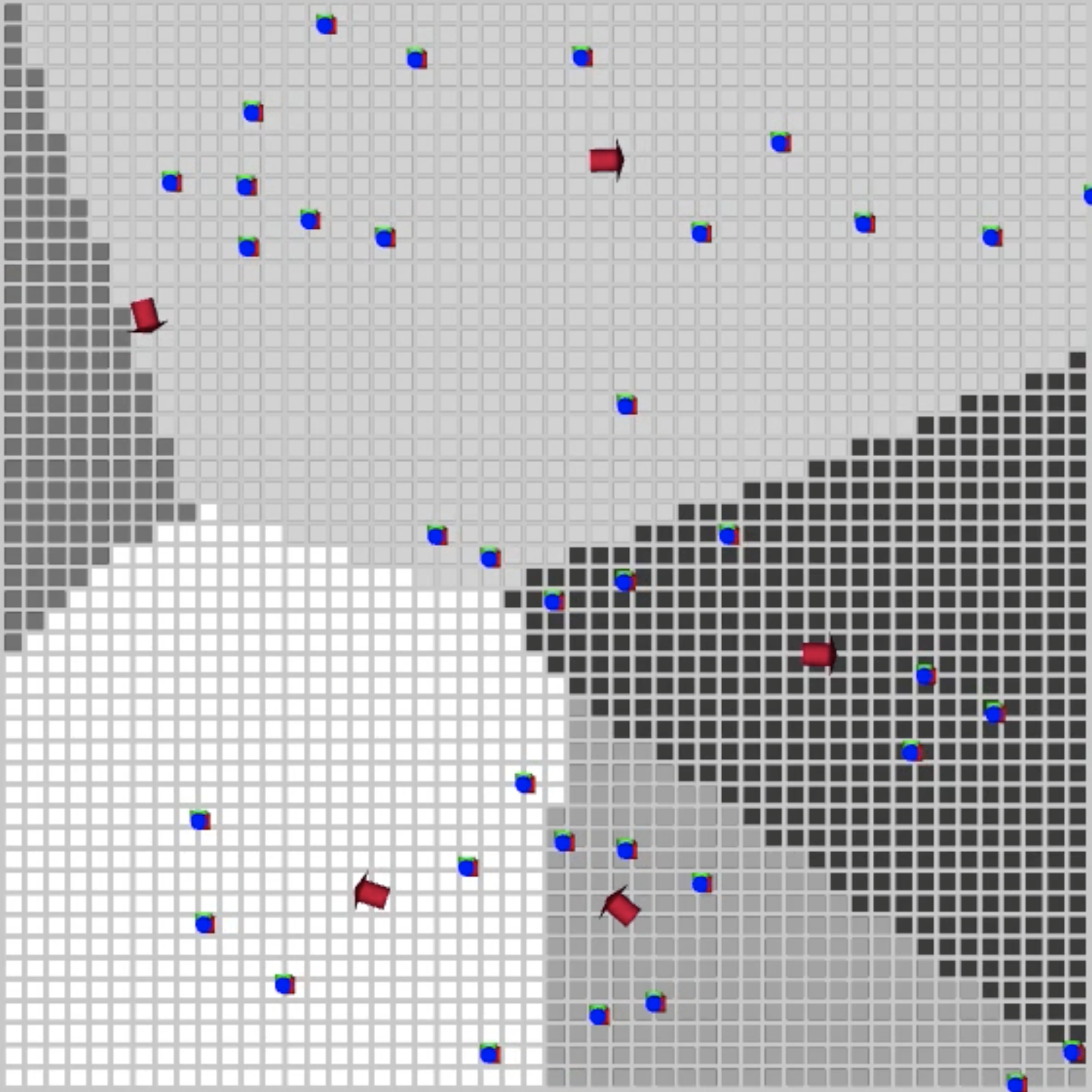}
	\label{fig:rviz_520}
}  \hspace{0.4cm}
\subfloat[RViz 6'30'']{
	\includegraphics[width=0.5\columnwidth]{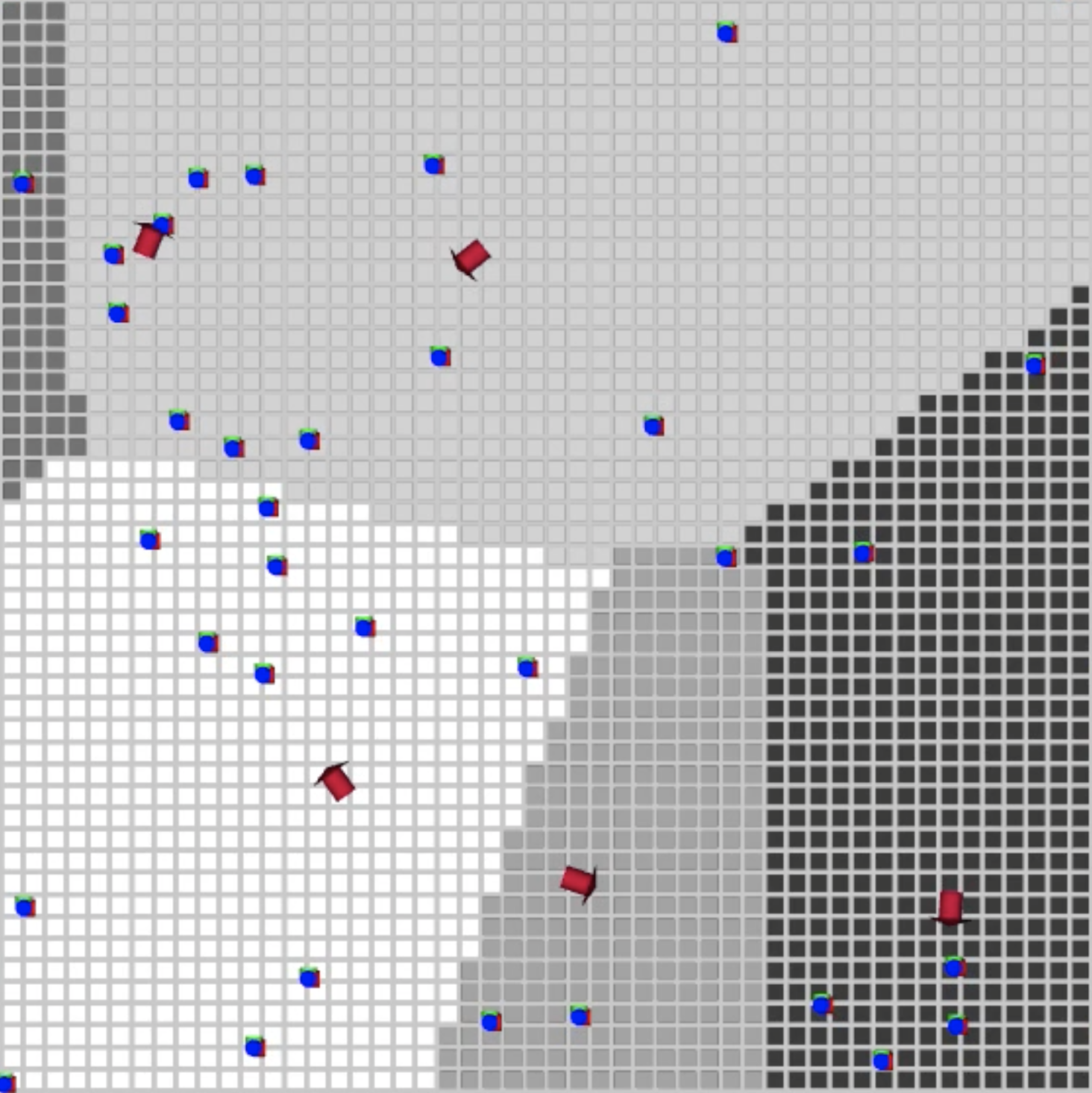}
	\label{fig:rviz_630}
}
\caption{Figures showing four clips during a single test using five TurtleBot3 robots, one of each type in Table~\ref{table:tb3_sensors}, to track forty moving targets, \ie from starting moment to \unit[2]{min} \unit[30]{s} (Figures~\ref{fig:0_230},~\ref{fig:rviz_0},~\ref{fig:rviz_230}), from \unit[3]{min} \unit[40]{s} to \unit[4]{min} \unit[6]{s} (Figures~\ref{fig:340_406},~\ref{fig:rviz_340},~\ref{fig:rviz_430}), from \unit[4]{min} \unit[30]{s} to \unit[5]{min} \unit[12]{s} (Figures~\ref{fig:430_512},~\ref{fig:rviz_430},~\ref{fig:rviz_520}), and from \unit[5]{min} \unit[20]{s} to \unit[6]{min} \unit[30]{s} (Figures~\ref{fig:520_630},~\ref{fig:rviz_520},~\ref{fig:rviz_630}), respectively. Figures on the first \edit{column} are screenshot overlays of Gazebo GUI \edit{taken at the final time in the time interval}, showing the top-view of five robots with FoVs, marked in red, and targets marked in \edit{magenta} dots. Robot trajectories and target traces are also shown. Figures on the second and the third \edit{column} show screenshot of RViz GUI at the beginning and the end of each clip, respectively. Red arrows show robot locations and orientations. Blue dots show target locations. Regions in different shades of grey show the assigned cells for each robot.}
\label{fig:track}
\end{figure*}

The robots begin with a uniform PHD where we set the expected number of targets equal to $1$ \edit{over the entire task space}, meaning that they have no prior knowledge of the target density distribution. 
At the beginning of the simulations, the robots are located alongside the lower edge of the task space where they cannot detect the majority of the targets.
Four animation clips excerpted from a 12 minute 30 second long target tracking simulation are presented in Figure~\ref{fig:track}. 
Each column corresponds to one of the clips illustrating the following three \edit{behaviors exhibited by the robot network}.

\begin{figure}[ht]
\centering
\includegraphics[width=0.65\columnwidth]{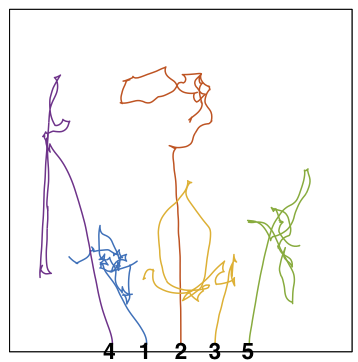}
\caption{Figure showing trajectories over \unit[12]{min} \unit[30]{s} testing time of \edit{5 robots in the square open task space}.The numbers indicate the IDs of the robots.}
\label{fig:gazebo_traj}
\end{figure}
\begin{figure}[ht]
\centering
\includegraphics[width=0.9\columnwidth]{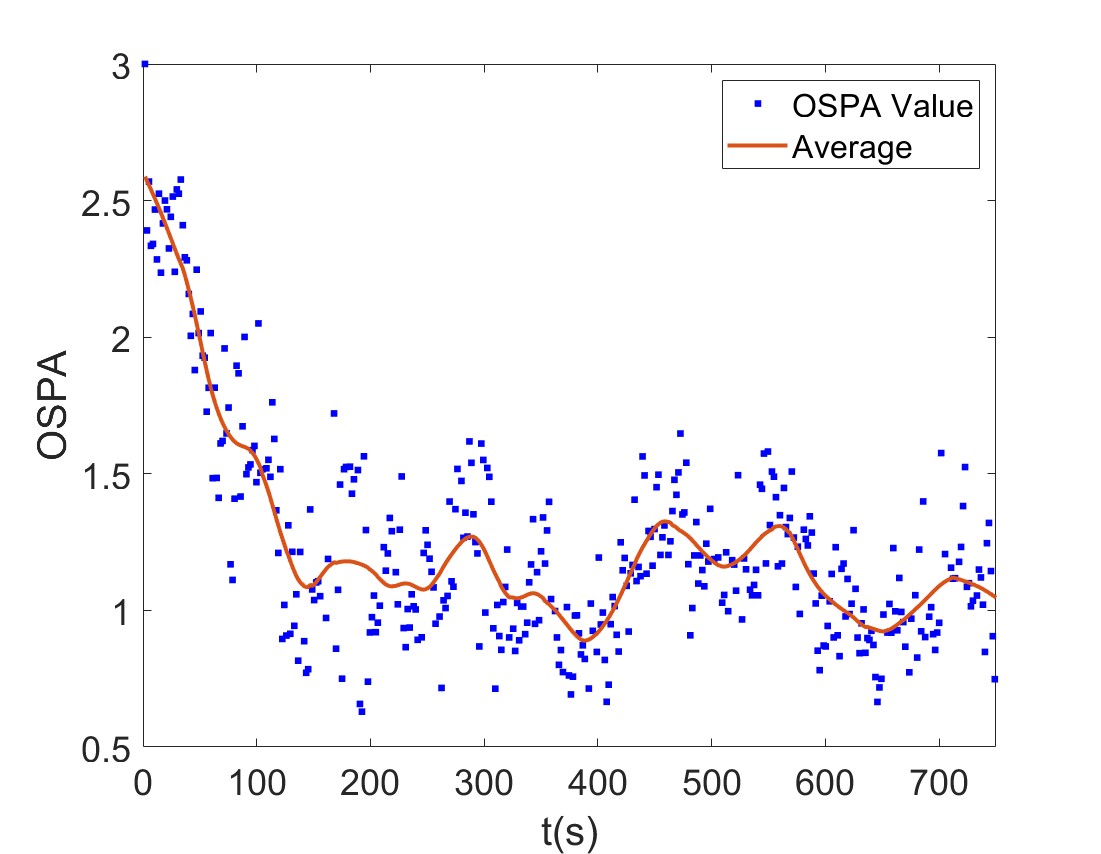}
\caption{Figure showing OSPA error at each discrete time step and its moving average \edit{over the previous five data points} during the entire simulation time.}
\label{fig:gazebo_ospa}
\end{figure}
\begin{figure*}[ht]
\centering
\subfloat[Robot 1]{
	\includegraphics[width=0.65\columnwidth]{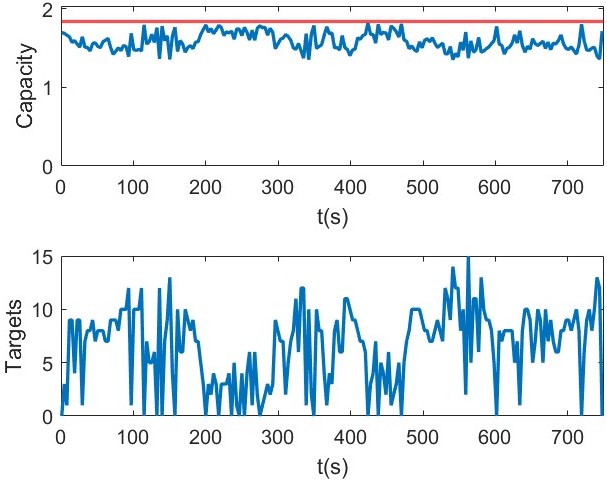}
	\label{fig:cap1}
}
\subfloat[Robot 2]{
	\includegraphics[width=0.65\columnwidth]{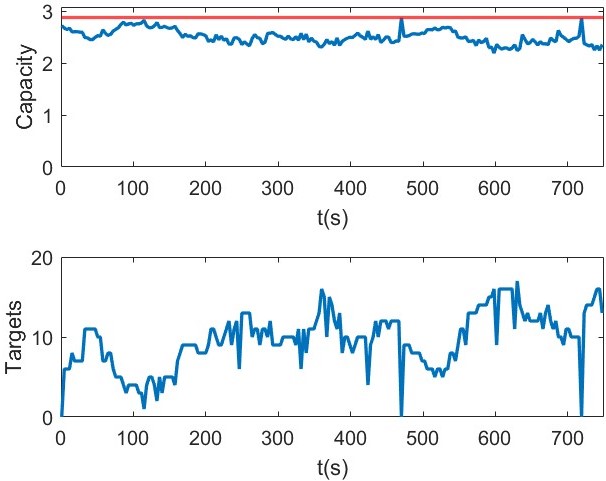}
	\label{fig:cap2}
}
\subfloat[Robot 3]{
	\includegraphics[width=0.65\columnwidth]{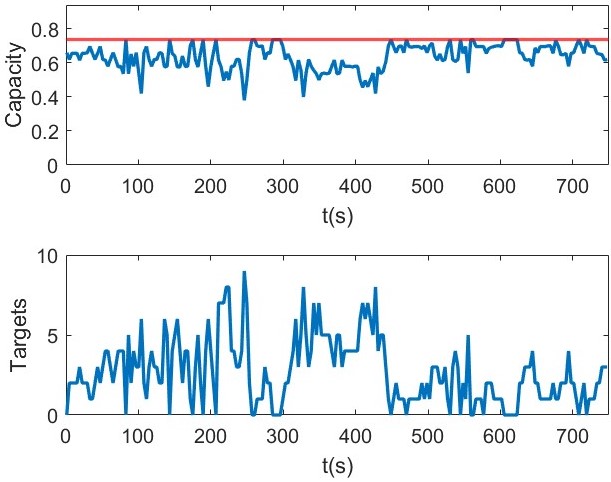}
	\label{fig:cap3}
} \\
\subfloat[Robot 4]{
	\includegraphics[width=0.65\columnwidth]{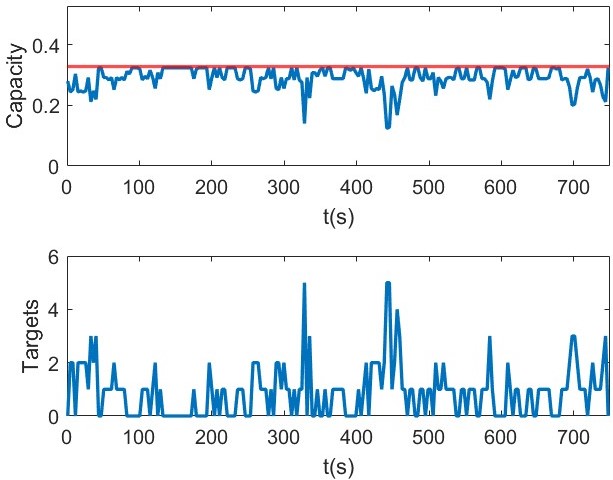}
	\label{fig:cap4}
} 
\subfloat[Robot 5]{
	\includegraphics[width=0.65\columnwidth]{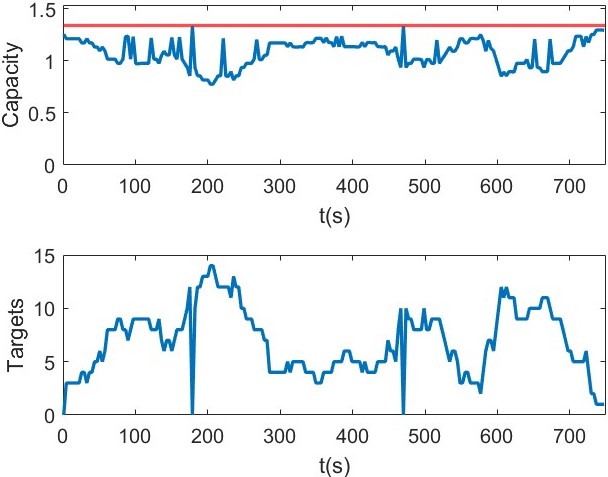}
	\label{fig:cap5}
}
\subfloat[Total]{
	\includegraphics[width=0.65\columnwidth]{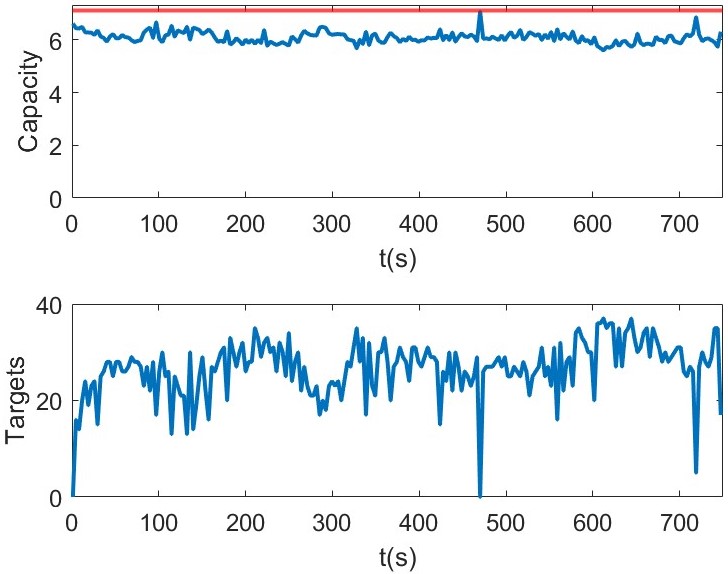}
	\label{fig:cap_total}
}
\caption{Figures plot the normalized unused sensing capacities $U$ (abbreviated as "Capacity" for the y-axis) and the numbers of detected targets for all five robots \edit{and their summation over the entire testing time}. Red horizontal lines show the maximum sensing capacities $C_{max}$.}
\label{fig:capacity vs target}
\end{figure*}
\begin{figure}[ht]
\centering
\subfloat[4'26'']{
	\includegraphics[width=0.67\columnwidth]{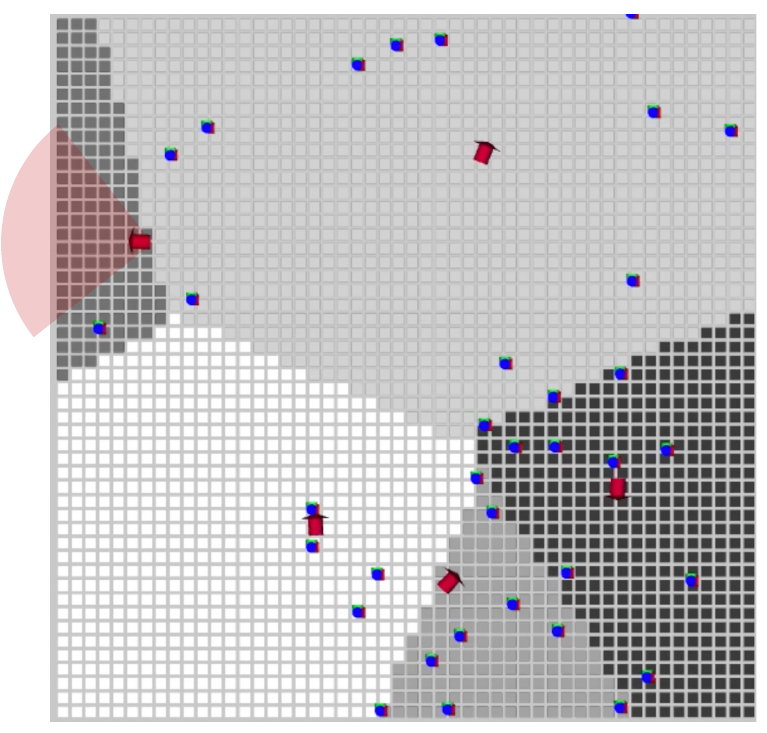}
	\label{fig:30t}
} \\
\subfloat[\edit{4'28''}]{
	\includegraphics[width=0.66\columnwidth]{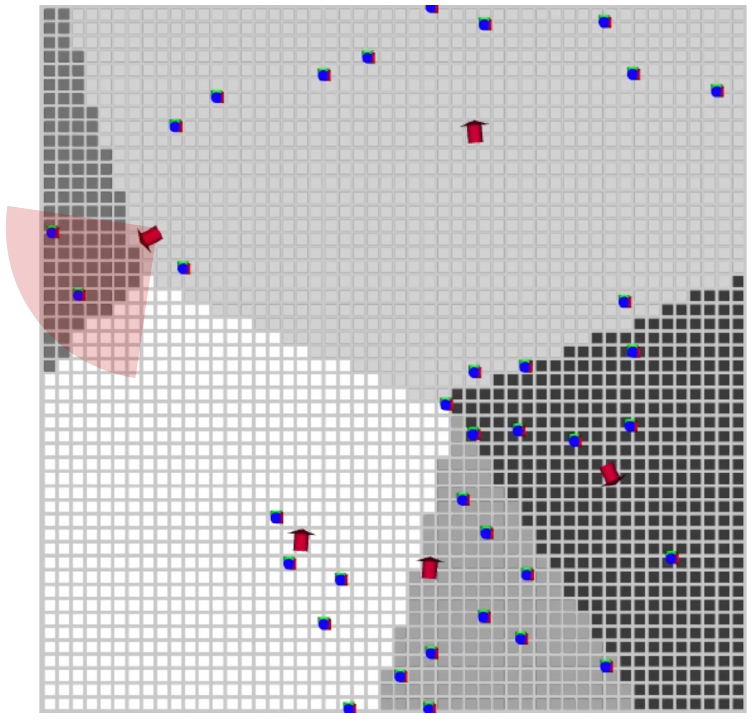}
	\label{fig:40t}
}
\caption{Figures showing the number of cells assigned to the top-left robot switching from 140 at \unit[4]{min} \unit[26]{s} to 109 after 2 seconds.}
\label{fig:comparison}
\end{figure}

\paragraph{Target Coverage}
Our algorithm guides the robot to explore the task space when no targets are detected and to maintain coverage of the majority of detected targets when the robot detects multiple targets moving in different directions.
The first clip (\ie the first \edit{row} in Figure~\ref{fig:track}) shows the initial exploration process
where 5 robots start from their initial locations and move across the task space as the control input \eqref{eq:u_i} drives the COD of each robot to the centroid of its CCVD.

The area that each robot covers for target detection is continuously updated to respond to the targets' movement, and the sensor FoVs cover most of the areas with high target density, as observed in Figures~\ref{fig:340_406},\ref{fig:rviz_340},\ref{fig:rviz_406}, Figures~\ref{fig:430_512},\ref{fig:rviz_430},\ref{fig:rviz_512}, and Figures~\ref{fig:520_630},\ref{fig:rviz_520},\ref{fig:rviz_630}.
In the second clip, we can see that three robots in the bottom tend to gather close to each other as the targets densely aggregate in the lower-right corner of the task space while maintaining certain distance from one another to cover a larger area.
Across all figures, we can observe that a sensor with a higher sensing capability is assigned with a larger task region optimized by our task assignment method.

\paragraph{Target Following}
When a robot detects a small number of targets moving in about the same direction, the robot follows those targets. 
This can be observed in Figures~\ref{fig:520_630},\ref{fig:rviz_520},\ref{fig:rviz_630}, where the right-most robot follows three targets that are moving towards the bottom of the task space.

\paragraph{Target Tracking Reallocation}
When a robot detects multiple targets moving in different directions, it chooses to follow the targets that are within its assigned cells and reallocates the task of tracking other targets to its neighboring robots as the targets move into their cells.
This is illustrated in Figures~\ref{fig:430_512},\ref{fig:rviz_430},\ref{fig:rviz_512} where the right-most robot follows the multiple targets clustered at its top-left region as the weighted centroid of PHD over its assigned region shifts toward the target cluster.
As these targets move outside the robot's assigned region, they no longer dominate the weighted centroid, causing the robot to follow the other three targets on its right to maintain coverage of the bottom-right corner of the task space, shown in the fourth clip.
Such responsive switching behavior in the target coverage and following allows the robots to effectively gather information on the target's locations.

Robot trajectories over the entire simulation time are plotted in Figure~\ref{fig:gazebo_traj}.
Each robot first spreads out to deploy across the task space, indicated by the relatively straight part of trajectories.
Then each robot moves twistingly as it searches for or tracks targets within a relatively smaller area of the task space.
The trajectories reveal that robots with greater sensing capabilities, \eg robot 1 and 2, move within a smaller range after initial deployment, while those with weaker sensing capabilities, \eg robot 3, 4, and 5, move within a wider range.
This is due to the fact that the former robots often detect more targets moving in various directions and perform target coverage behavior within a relatively fixed region, while the latter ones more frequently detect fewer targets and perform target following across a wider sub-region.
Figure~\ref{fig:gazebo_ospa} plots the OSPA error with $p=1$ and $c=3$ in \eqref{eq:ospa}, which indicates that targets are effectively tracked after the initial deployment of the robotic network as the OSPA error significantly decreases comparing to the initial state.

Figure~\ref{fig:capacity vs target} plots the changes in the \edit{robots' normalized unused sensing capacity} as the numbers of detected targets changes over time. 
Overall, the normalized unused sensing capacity tends to decrease as the number of detected targets increases for each robot, reflecting the online adaptability of task assignment to the workload.
However, such increase and decrease do not strictly correspond to each other due to the presence of sensor false alarms (\ie false negative or false positive detections).

Figure~\ref{fig:comparison} visualizes the variation in the number of cells assigned to the top-left robot as the number of detected targets changes.
As the robot detects more targets,its normalized unused sensing capacity is reduced and, hence, so does its assigned task region.
Consequently, a large coverage region is assigned to its neighbors \edit{,} whose normalized unused sensing capacities are unchanged.

\subsection{Quantitative Results}
\label{subsec:quantitative}
\begin{table}[tbp]
\centering
\caption{\edit{Types} of Heterogeneous Sensors}
\label{table:type}
\begin{tabular}{| c || c | c | c | c | c |}
\hline
\backslashbox{Types}{Specs} & \thead{$\Theta_i$\\ (deg)} & \thead{$L_i$ \\(\unit{m}) } & $f_{d,i}(\Delta L)$ & $C_{max}$\\
\hline
\hline
\textrm{A} & 45 & 8 & 0.99 & 24.88\\ 
\hline
\textrm{B} & 45 & 8 & 0.7 & 17.59\\
\hline
\textrm{C} & 240 & 8 & 0.99 & 134.04\\
\hline
\textrm{D} & 270 & 11.3 & 0.99 & 300.86\\
\hline
\textrm{E} & 270 & 16 & 0.99 & 603.17\\
\hline
\end{tabular}
\end{table}
\begin{table}[tbp]
\centering
\caption{Network Compositions}
\label{table:comp}
\begin{tabular}{| c || c | c | c | c | c || c | c |}
\hline
\backslashbox{Comp}{\edit{Type}} & \textrm{A} & \textrm{B} & \textrm{C} & \textrm{D} & \textrm{E} & $C(S)$ & $L(S)$\\
\hline
\hline
$S_1$ & 6 & 18 & 12 & - & - & 2074.4 & \edit{3.7}\\
\hline
$S_2$ & 8 & 24 & 16 & - & - & 2765.8 & \edit{3.7}\\ 
\hline
$S_3$ & 10 & 30 & 20 & - & - & 3457.3 & \edit{3.7}\\
\hline
$S_4$ & 16 & - & - & - & 2 & 1604.4 & 6.1\\ 
\hline
$S_5$ & 16 & - & - & 4 & - & 1604.4 & 4.9\\
\hline
$S_6$ & 16 & - & 9 & - & - & 1604.4 & 3.1\\
\hline
\end{tabular}
\end{table}
\begin{figure}[tbp]
\centering
\includegraphics[width=0.75\columnwidth]{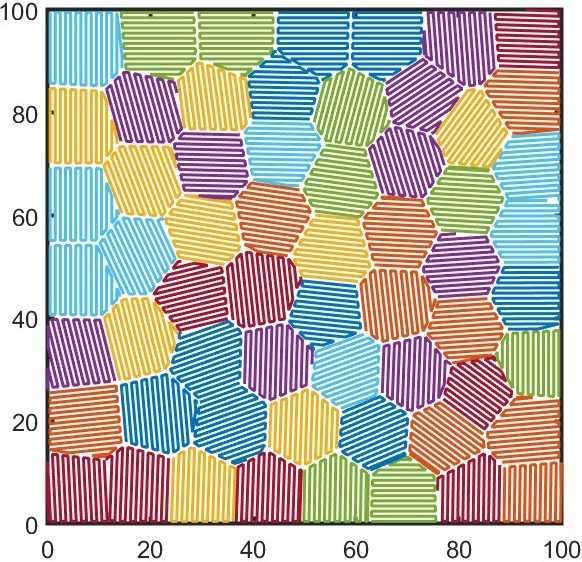}
\caption{\edit{An example of coverage paths for a sixty-robot team. Colored curves show planed coverage paths for each robot to cover its assigned Voronoi cell. Each robot must move to the starting point of its coverage path first from its original location at the beginning of each simulation trial.}}
\label{fig:zigzag}
\end{figure}
\begin{figure*}[tbp]
\centering
\subfloat[\edit{36 Robots (S1)}]{
	\includegraphics[width=0.65\columnwidth]{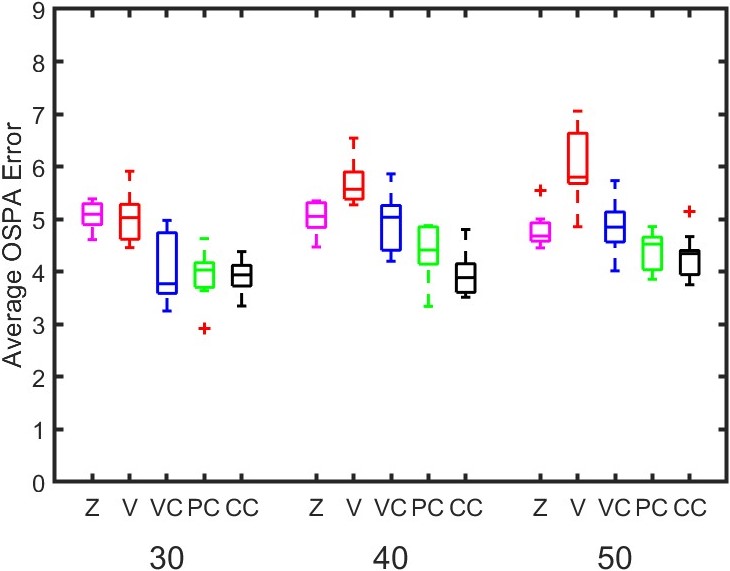}
	\label{fig:r36}
}
\subfloat[\edit{48 Robots (S2)}]{
	\includegraphics[width=0.65\columnwidth]{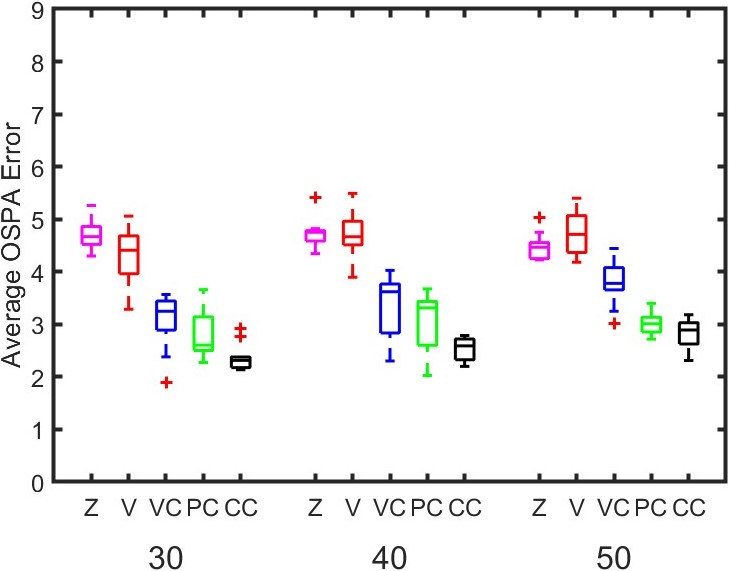}
	\label{fig:r48}
}
\subfloat[\edit{60 Robots (S3)}]{
	\includegraphics[width=0.65\columnwidth]{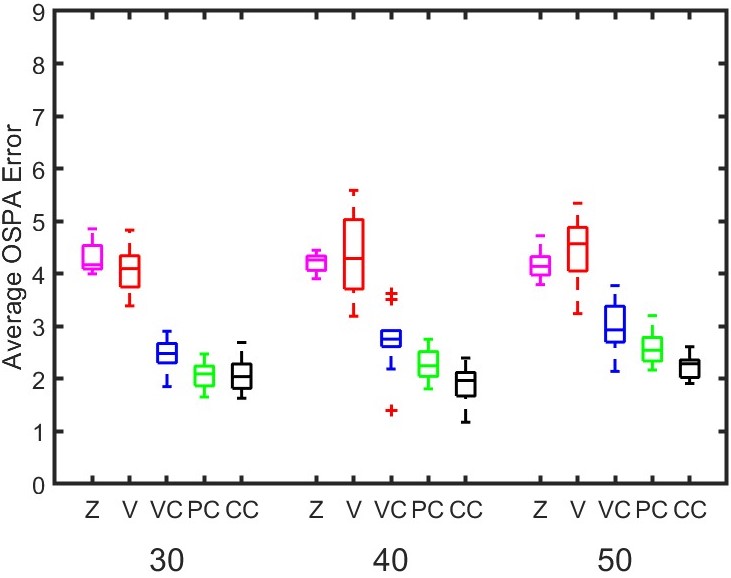}
	\label{fig:r60}
}
\caption{Boxplots showing the \edit{median and the 25th and 75th percentiles of} average OSPA error for each test configuration. \edit{Blue, red, magenta, green, and black boxplots show results of Zigzag (Z) method, Voronoi (V) method, Voronoi-COD (VC) method, Power-COD (PC) method, and CCVD-COD (CC) method, respectively. } Figure \ref{fig:r36}, \ref{fig:r48} and \ref{fig:r60} show results of network size \edit{36 ($S_1$), 48 ($S_2$), and 60 ($S_3$)} respectively. Each boxplot show results of a robotic network with certain size tracking 30, 40, and 50 moving targets respectively from left to right.}
\label{fig:ospa_comparison}
\end{figure*}
\begin{table*}[ht]
\centering
\caption{P-value of Average OSPA}
\label{table:p_values}
\begin{tabular}{| c || c | c | c || c | c | c || c | c | c |}
\hline
Team &
  \multicolumn{3}{c ||}{36 Robots (S1)} &
  \multicolumn{3}{c ||}{48 Robots (S2)} &
  \multicolumn{3}{c |}{60 Robots (S3)} \\
\hline
\hline
\backslashbox{Methods}{Targets} & 30 Targets & 40 Targets & 50 Targets & 30 Targets & 40 Targets & 50 Targets & 30 Targets & 40 Targets & 50 Targets \\
\hline
\hline
VC-PC & 0.8468 & \textbf{0.0783} & 0.1128 & 0.2499 & 0.2477 & \textbf{0.0001} & \textbf{0.0155} & \textbf{0.0585} & \textbf{0.0731} \\
\hline
PC-CC & 0.9209 & \textbf{0.0605} & 0.5158 & \textbf{0.0225} & \textbf{0.0194} & \textbf{0.0998} & 0.8474 & \textbf{0.0283} & \textbf{0.0137} \\ 
\hline
VC-CC & 0.7695 & \textbf{0.0010} & \textbf{0.0433} & \textbf{0.0021} & \textbf{0.0007} & \textbf{0.0000} & \textbf{0.0162} & \textbf{0.0020} & \textbf{0.0007} \\
\hline
\end{tabular}
\end{table*}
\begin{figure*}[tbp]
\centering
\subfloat[Voronoi Diagram]{
	\includegraphics[width=0.65\columnwidth]{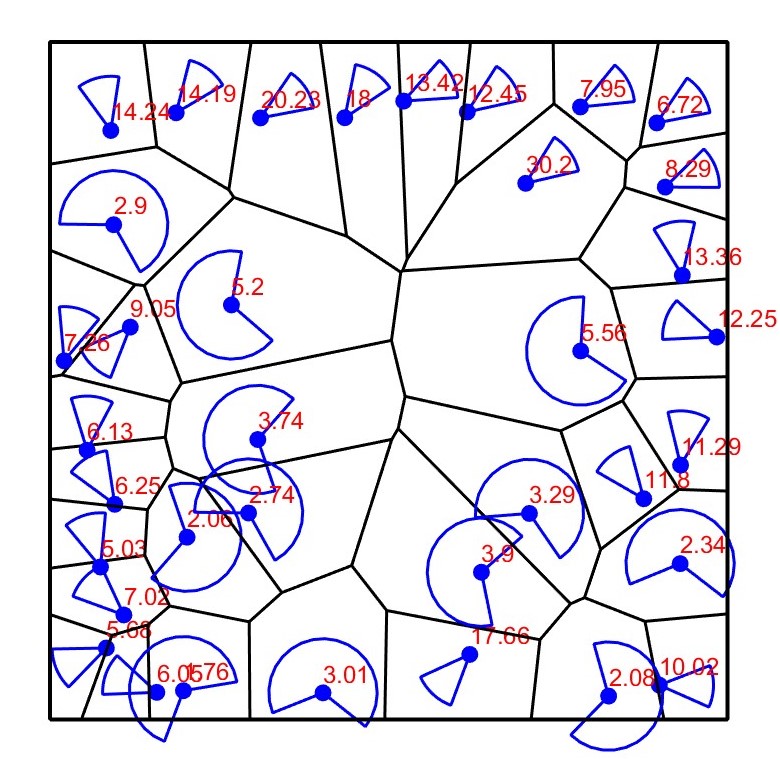}
	\label{fig:voronoi_36r}
}
\subfloat[Power Diagram]{
	\includegraphics[width=0.65\columnwidth]{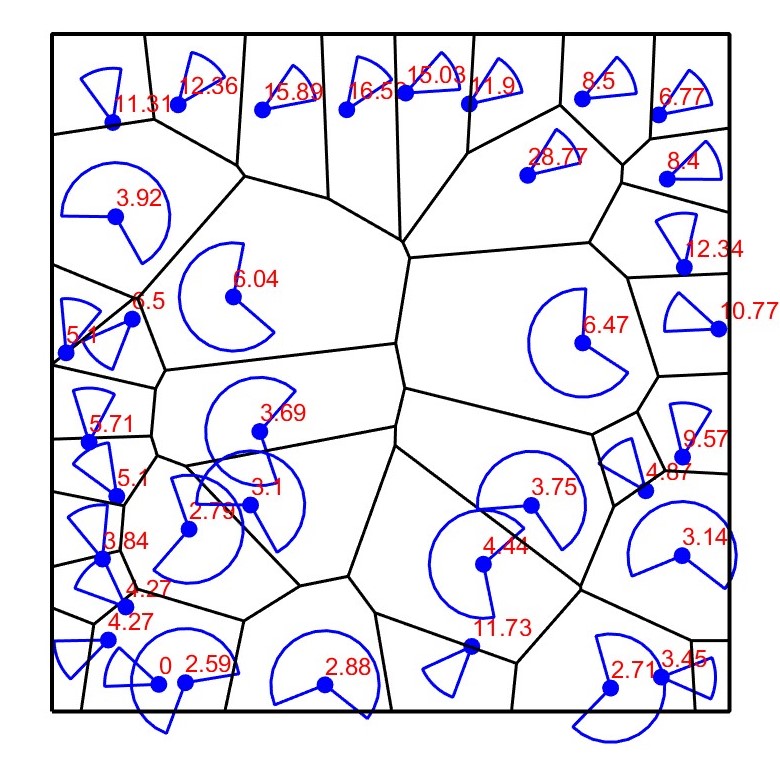}
	\label{fig:power_36r}
} 
\subfloat[CCVD]{
	\includegraphics[width=0.65\columnwidth]{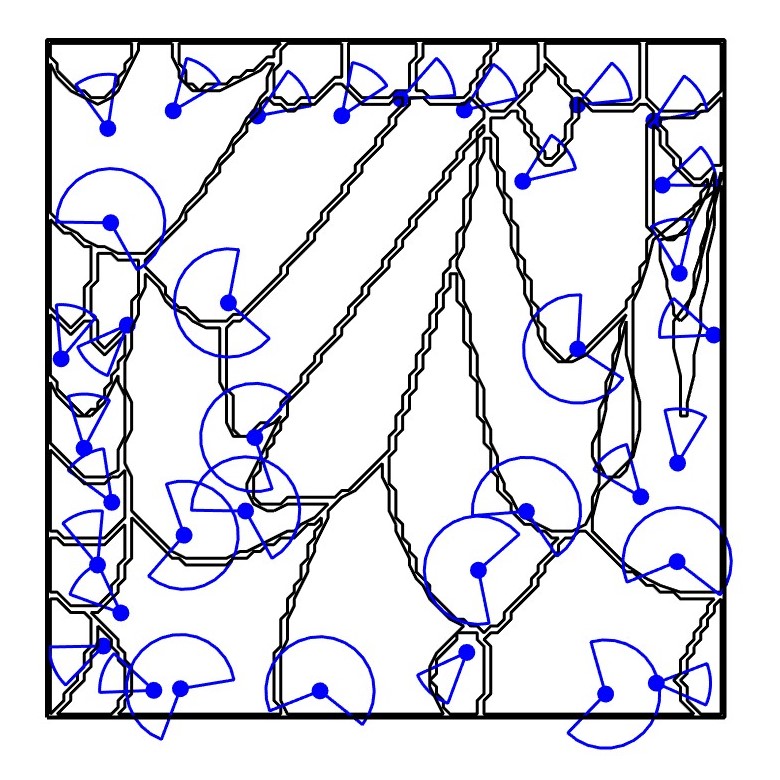}
	\label{fig:ccvd_36r}
}
\caption{\edit{Figures show an instance of a simulation trial using 36 robots. Blue circles and sectors plot robot locations and footprints. Black boundaries plot space partitions. In Figures~\ref{fig:voronoi_36r} and~\ref{fig:power_36r}, red numbers indicate the area-to-capacity ratio of each robot. Note that the serrated boundaries in Figure~\ref{fig:ccvd_36r} are due to the discrete implementation of CCVD construction.}}
\label{fig:instance_36r}
\end{figure*}
\begin{figure}[tbp]
\centering
\includegraphics[width=0.7\columnwidth]{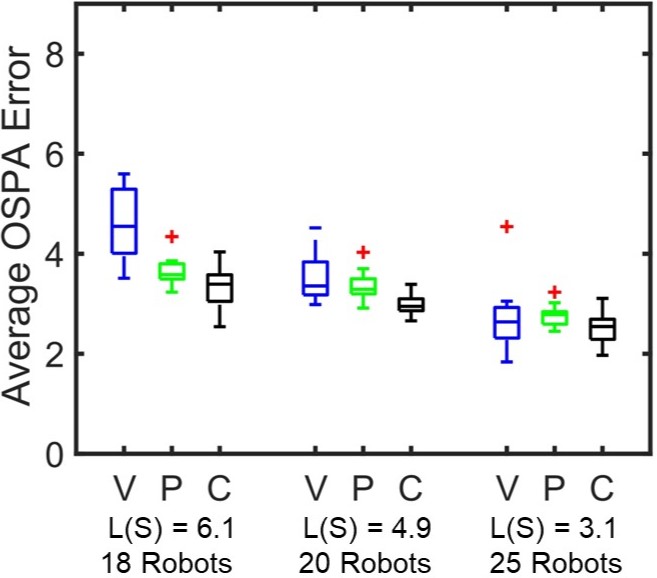}
\caption{Boxplot shows the OSPA errors under heterogeneity levels 1-3 from left to right. \edit{Blue, green, and black boxplots show the results of using the Voronoi diagram (V), power diagram (P), and CCVD (C), respectively.}}
\label{fig:levels}
\end{figure}

To quantitatively compare the efficacy of our \edit{power diagram (Section~\ref{subsec:power diagram}) and CCVD (Section~\ref{subsec:ccvd}) based method with baseline algorithms}, we run batches of simulation trials in \textsc{MATLAB} using a point robot model with dynamic model given as \eqref{eq:dynamic}.
\edit{In order to demonstrate the effectiveness of two novel algorithms and to perform ablation experiments, we apply five methods in the following tests:
\begin{enumerate}
  \item Zigzag (Z) Method: This method utilizes a complete coverage path planning framework, a standard approach for target search. Firstly, a centroidal Voronoi diagram of $n$ Voronoi sites is computed using Lloyd's algorithm. Each robot is assigned a Voronoi cell of the diagram. Then, a zigzag coverage path is planned \cite{torres2016coverage} with the spacing of two adjacent parallel paths equaling to \unit[1]{m} for each robot to exploit its cell. An example of paths planned for 60 robots is shown in Figure~\ref{fig:zigzag}. Each robot tracks targets using PHD filter while traversing through the pre-planned coverage path.
  \item Voronoi (V) Method: Partitioning the space using Voronoi diagram with robot locations as generator points, similar to Dames' method \cite{dames2020distributed}.
  \item Voronoi-COD (VC) Method: Similar to Voronoi method besides using robot CODs as generator points.
  \item Power-COD (PC) Method: Our first proposed method, \ie partitioning the space using power diagram with robot CODs as generator points.
  \item CCVD-COD (CC) Method: Our second proposed method, \ie partitioning the space using CCVD with robot CODs as generator points.
\end{enumerate} }
In these simulations, the size of the environment $E$ is \edit{$\unit[100]{m} \times \unit[100]{m}$}.
Robots move at a maximum linear velocity of $\unit[2]{m/s}$ and a maximum angular velocity of $\unit[2]{rad/s}$.
Each robot carries one of the five sensor \edit{types}, as outlined in Table~\ref{table:type}.
Assuming that the target density within the FoVs of sensors are completely unknown, we select $\mu = 1$ in \eqref{eq:det_max} for all robots in all trials.
Based on these \edit{type} definitions, we define six different network compositions, shown in Table~\ref{table:comp} \edit{in which columns A-E list the number of robots with each sensor type}.
Distinguished from Section~\ref{subsec:qualitative}, in this series of simulations, target motion has a higher degree of randomness.
Targets move at a maximum velocity of $\unit[1]{m/s}$ with their heading directions randomly changing over time, and may enter or leave the environment by crossing over the boundary.
This means that the number of targets varies over time, and the true number of targets is not known to the sensors.

\subsubsection{Team Size}
Initially, there are 30, 40, or 50 targets within the environment.
We compare three different network compositions, $S_1$, $S_2$, and $S_3$, each of which is composed of the same sensor types in equal proportions, as shown in Table~\ref{table:comp}. 
By varying the number of targets and the network compositions, we create 9 different simulation trials. 
Each trial is repeated for 10 times using the three algorithms, each lasting for $\unit[700]{s}$.
The robots and targets are randomly located in the task space at the beginning of each trial.

Figure \ref{fig:ospa_comparison} shows the results of the nine trials.
We plot \edit{the median and the 25th and 75th percentiles of average OSPA via boxplots} using the data collected during the last $\unit[400]{s}$ of each trial to present the steady state tracking performance.
Note that larger teams performs better in the target tracking.
\edit{We also report P-values for comparison between methods VC and PC, PC and CC, VC and CC, respectively, and bold the data that are significant at the 10\% level of significance.}
Consistently, the CCVD methods (CC) outperform the power diagram (PC) and Voronoi diagram methods (V and VC) across different team sizes and target numbers. 
This confirms the improved efficacy of our proposed CCVD methods, especially for heterogeneous sensing networks.
The CCVD and Power diagrams assign larger regions to the sensors with higher unused sensing capacity and smaller regions to those who have smaller unused sensing capacity.
The CCVD places a hard constraint, \ie the capacity constraint, on the size of the partition depending on each robot's workload. 
Hence it demonstrates a further improvement in the tracking performance over the power diagram, which instead utilizes an approximation scheme which associates the unused sensing capacity of a robot with a power radius in space partition.
As a result, sensing networks can take advantage of their heterogeneous sensing capabilities and avoid overloading sensors with target tracking tasks.

\edit{Figures~\ref{fig:instance_36r} visualize a random moment in a simulation trial with 36 robots, which is utilized as a case study to demonstrate the improvement in space partition optimality. 
For each robot, we compute the value of area of its assigned space divided by its normalized unused sensing capacity, which reveals the among of a robot's workload comparing to its sensing capability, and plot it on Figure~\ref{fig:power_36r}.
It is shown that the area-to-capacity ratio of both Voronoi method and power method varies substantially among robots, indicating that robots are more or less assigned a task area size that does not match their capabilities.
PC method improves the ratio to a standard deviation of 5.5, comparing with that of VC method, which is 6.2.
However, CC method yields a constant area-to-capacity ratio over the team at all time to further optimize the space partition.}

\edit{Moreover, it can be seen that VC method consistently yields to significant lower OSPA error than V method, and that PC method significantly ourperforms VC method in most of the simulation trials.
This reveals that both replacing robot locations to CODs as generator points and applying the capacity-constraint space partitioning strategies contribute to the tracking accuracy, leading to the out-performance of PC and CC methods.
We also find that PC and CC methods significantly improve the tracking accuracy comparing with Z method, indicating that our proposed algorithms produce more effective multi-robot path planning results for heterogeneous teams to track multiple unknown moving targets. 
}

In addition to decreasing the average tracking error, \edit{our proposed algorithms, especially CCVD, also decreases the range of \edit{OSPA, \ie} tracking errors in a majority of cases, revealed from Figure~\ref{fig:ospa_comparison} and Figure~\ref{fig:levels} introduced in Section~\ref{subsec:levels}}.
\edit{In Table~\ref{table:std}, we list the standard deviation of OSPA over all trials in Figure~\ref{fig:ospa_comparison}}.
In other words, sensing networks that use the CCVD or power diagram often perform more reliable behavior.
This is due to the fact that the optimized assignment of dominance region reduces the probability of targets being lost during tracking.

\edit{\begin{remark}[Comparison of Results]
    The three partitioning methods (Voronoi diagrams, power diagrams, and the CCVD) each use a different function $f(\cdot)$ in the functional $\mathcal{H}$ defined in Equation~\eqref{eq:optimization}. We see in Figure~\ref{fig:diagrams} that the three methods can lead to quite different partitions, with Voronoi and power diagrams tending to look more similar to one another. Given this, it is an interesting result that power and CCVD both yield comparable results in terms of tracking accuracy. We believe that these results emphasize the utility of considering an equitable distribution of space based on each robot's capability and operational conditions, regardless of the specific partitioning method employed.
\end{remark}}

\begin{table}[tbp]
\centering
\caption{Standard Deviation of OSPA}
\label{table:std}
  \begin{tabular}{ p{1.15cm} || p{0.67cm} | p{0.67cm} || p{0.67cm} | p{0.67cm} || p{0.67cm} | p{0.67cm} }
    \hline
    Target &
      \multicolumn{2}{c ||}{30 Targets} &
      \multicolumn{2}{c ||}{40 Targets} &
      \multicolumn{2}{c}{50 Targets} \\
    \hline
    Method & CC & V & CC & V & CC & V \\
    \hline
    \hline
    36 Robots & \textbf{0.3167} & 0.4761 & \textbf{0.4212} & 0.4353 & \textbf{0.4097} & 0.7183\\ 
    \hline
    48 Robots & \textbf{0.2488} & 0.6019 & \textbf{0.2166} & 0.5192 & \textbf{0.2643} & 0.4075\\
    \hline
    60 Robots & \textbf{0.3378} & 0.4593 & \textbf{0.3774} & 0.8131 & \textbf{0.1951} & 0.5976\\
    \hline
  \end{tabular}
\end{table}

\subsubsection{Heterogeneity Levels}
\label{subsec:levels}
Lastly we run a series of tests to validate the following observation.
\edit{For a team with a fixed total sensing capacity, $C(S)$, our novel CCVD and power diagram formula will provide increasing improvements in target tracking accuracy over a standard Voronoi diagram as the heterogeneity, $L(S)$, increases.}

We use compositions 4, 5, and 6 from Table~\ref{table:comp}, all of which have the identical total sensing capacity of 1604.4 determined by \eqref{eq:total}. 
There are 30 targets in all simulation trials.
The heterogeneity levels \eqref{eq:level} for the three network are given as $L(S_4) = \unit[6.1]{m}$, $L(S_5) = \unit[4.9]{m}$, and $L(S_6) = \unit[3.1]{m}$, respectively.

Figure~\ref{fig:levels} shows the average OSPA errors over 10 runs, using data from the last \unit[400]{s} of each trial.
We observed significant improvement in the tracking performance of the power diagram and CCVD method compared to that of the standard Voronoi diagram when the network is highly heterogeneous \edit{($S_4$)}.
On the other hand, as the heterogeneity level decreases, so does the performance improvement as we can observe from $S_5$ and $S_6$.
These results support our \edit{aforementioned observation}.
\edit{If all sensors are the same, the power distance in Equation~\eqref{eq:power distance} will only depend on sensor's location so that the locational optimization functional \eqref{eq:optimization} will converge to the one using Voronoi diagram. 
Meanwhile, since the capacity of the cells tends to be close to each other, the CCVD will get close to the Voronoi diagram as well.
We also find that the improvement in tracking accuracy by $S4$, $S5$ and $S6$ increases with the number of robots in the network}, which affects the mobility for a network.

\edit{\begin{remark}[Trade-off between algorithms]
    The \textsc{MATLAB} simulations were conducted on a Windows 11 laptop with 13th Gen Intel(R) Core(TM) i7-1360P 2.20GHz processor and 32GB RAM storage memory. 
    Taking the simulations of 60 robots as an example, the average running time of Algorithm~\ref{alg:control} using PC method is approximately \unit[0.1]{s} per iteration, while running time of using CC method is around \unit[9.0]{s} per iteration.
    At the expense of running time, CC method reduces OSPA error by around 20\% comparing with PC method.
    Practitioners should make a trade-off between the two algorithms according to the availability of computing resources and the requirement of tracking accuracy.
\end{remark}}

\section{Conclusions}
\label{sec:conclusions}
We propose a distributed coverage control scheme for heterogeneous mobile robots with onboard sensors to track an unknown and time-varying number of targets.
This novel strategy allows sensors to have arbitrary sensing models (with limited \edit{fields of view}) and dynamically optimizes the workload for each individual.
To do this, we introduce the normalized unused sensing capacity to quantify the instant detection capability of each sensor.
We then use this to construct either a power diagram or a capacity constraint Voronoi diagram to recursively find optimized sensor locations as measurements are updated.
The centroid of detection for each sensor is utilized as the generation point to create the partitions, allowing each sensor to center its field of view on the area with the highest information density.
Simulation results show the convergence of our proposed method in target tracking scenarios and indicates that our method yields better and more reliable tracking performance compared to \edit{baseline algorithms} that does not account for heterogeneity.
We also observed that the tracking performance of our approach increases \edit{over the Voronoi approach} as the heterogeneity in the team's sensing capabilities increases.

\bibliographystyle{IEEEtran}
\bibliography{bib/root}


 




\vfill

\end{document}